\documentclass[11pt]{article} %
\usepackage[final]{acl}
\usepackage{times}
\usepackage{latexsym}
\usepackage{amsmath}
\usepackage{amssymb}
\usepackage{graphicx}
\usepackage{booktabs}
\usepackage[T1]{fontenc}
\usepackage[utf8]{inputenc}
\usepackage{xcolor}
\usepackage[most]{tcolorbox}
\tcbuselibrary{skins, breakable}
\definecolor{c-green-light}{RGB}{221,247,240}
\definecolor{c-green-dark}{RGB}{6,105,77}
\newtcolorbox{findingbox}[1]{%
  colback=white,
  colframe=c-green-dark,
  enhanced,
  coltitle=black,
  colbacktitle=c-green-light,
  title={\small\textcolor{c-green-dark}{\textbf{#1}}},
  boxsep=2.7pt,
  top=1pt,
  bottom=1pt,
  left=4pt,
  right=4pt}
\newtcolorbox{promptbox}[1]{%
  colback=gray!3,
  colframe=black!45,
  enhanced,
  breakable,
  coltitle=black,
  colbacktitle=gray!15,
  title={\small\textbf{#1}},
  boxsep=2.5pt,
  top=2pt,
  bottom=2pt,
  left=4pt,
  right=4pt,
  fontupper=\footnotesize,
  before upper={\raggedright}}

\usepackage{microtype}
\usepackage{hyperref}
\usepackage{url}
\usepackage{tabularx}
\usepackage{algorithm}
\usepackage{algpseudocode}
\usepackage{algcompatible}
\usepackage{graphicx}
\usepackage{wrapfig}
\usepackage{booktabs}
\usepackage{multirow}
\usepackage{array}
\newcolumntype{Y}{>{\raggedright\arraybackslash}X}
\usepackage[section]{placeins}
\graphicspath{{figures/}}

\usepackage{xspace}
\definecolor{darkblue}{rgb}{0, 0, 0.5}
\hypersetup{colorlinks=true, citecolor=darkblue, linkcolor=darkblue, urlcolor=darkblue}

\title{SearchAtlas: Analyzing Agentic Search Strategies via Evidential Query Graphs}

\author{%
  \textbf{Jiacheng Sang\textsuperscript{1}}\thanks{Equal contribution.} \quad
  \textbf{Mengyuan Li\textsuperscript{1}}\footnotemark[1] \quad
  \textbf{Sanxing Chen\textsuperscript{1}} \\
  \textbf{Yukun Huang\textsuperscript{1}} \quad
  \textbf{Yu Feng\textsuperscript{2}} \quad
  \textbf{Bhuwan Dhingra\textsuperscript{1}} \\
  \textsuperscript{1}Duke University \quad
  \textsuperscript{2}University of Pennsylvania \\
  \texttt{\{jiacheng.sang,alyssa.li,sanxing.chen,yukun.huang\}@duke.edu} \\
  \texttt{fengyu1@seas.upenn.edu} \quad
  \texttt{bdhingra@cs.duke.edu}
}

\newcommand{\qzero}{\ensuremath{q_0}\xspace}
\newcommand{\pk}{%
  \ifmmode\mbox{\textsf{Prior\_knowledge}}\else\textsf{Prior\_knowledge}\xspace\fi}
\newcommand{\qtag}[1]{%
  \ifmmode\mbox{\scriptsize\textsf{#1}}\else{\scriptsize\textsf{#1}}\fi}
\newif\ifshowedits
\showeditstrue

\begin{document}

\maketitle
\begin{abstract}
LLM search agents are often evaluated on final-answer accuracy, overlooking the process. Analyzing a search strategy requires understanding how credible evidence is retrieved to address question constraints. This valuable information is buried in raw search trajectories that are long and difficult to parse.
We introduce \textsc{SearchAtlas}, a framework that converts search trajectories into structured graphs whose edges represent how evidence is propagated across the reasoning trace, from the query that retrieves it to the final answer. Our automated parsing pipeline achieves a mean edge $F_1$ of $86.0\%$ against human-annotated graphs and remains consistent across repeated runs.
We analyze five search agents on three benchmarks, revealing systematic differences in search scale and evidence aggregation.
\textsc{SearchAtlas} exposes fragmented answer support, question constraints that do not reach the answer, and unverified parametric knowledge entering the response. These process failures are strongly associated with incorrect answers, even more so than an LLM judge given either the raw trajectory or the ordered query list, suggesting that the constructed graphs provide useful interpretability.
Moreover, an audit of cases in which process-diagnostic scores disagree with final-answer correctness shows that they capture information not reducible to answer accuracy.\footnote{Code and data are available at \url{https://github.com/DukeNLP/SearchAtlas}.}

\end{abstract}

\section{Introduction}
Search agents can plan multi-step web queries, visit pages, and synthesize answers over long horizons~\citep{li2025websailor,tongyi2025deepresearch,miromind2025mirothinker}.
A growing set of benchmarks evaluates these systems mostly on final-answer correctness~\citep{wei2025browsecomp,du2025deepresearchbench,xi2025infodeepseek,li2026deepresearchbench2,gupta2026deepsearchqa}, focusing on the outcome instead of the strategies of search agents. Two agents may differ drastically in how they approach an answer but obtain the same accuracy on a given benchmark.

The evidence needed to understand these strategies lies in search trajectories, which contain the agent's thoughts and actions, as well as feedback received from the environment. They are long and flattened into a chronological order of events, which obscures the information flow. For example, agents often adopt branched search strategies in which one piece of retrieved evidence informs several later queries. Conversely, multiple earlier results may be combined to support a single focused query. We therefore need a representation that makes the evidential dependencies explicit, allowing us to trace how an answer is supported and identify where the search process goes wrong.

\begin{figure*}[t]
  \centering
  \includegraphics[width=0.98\textwidth]{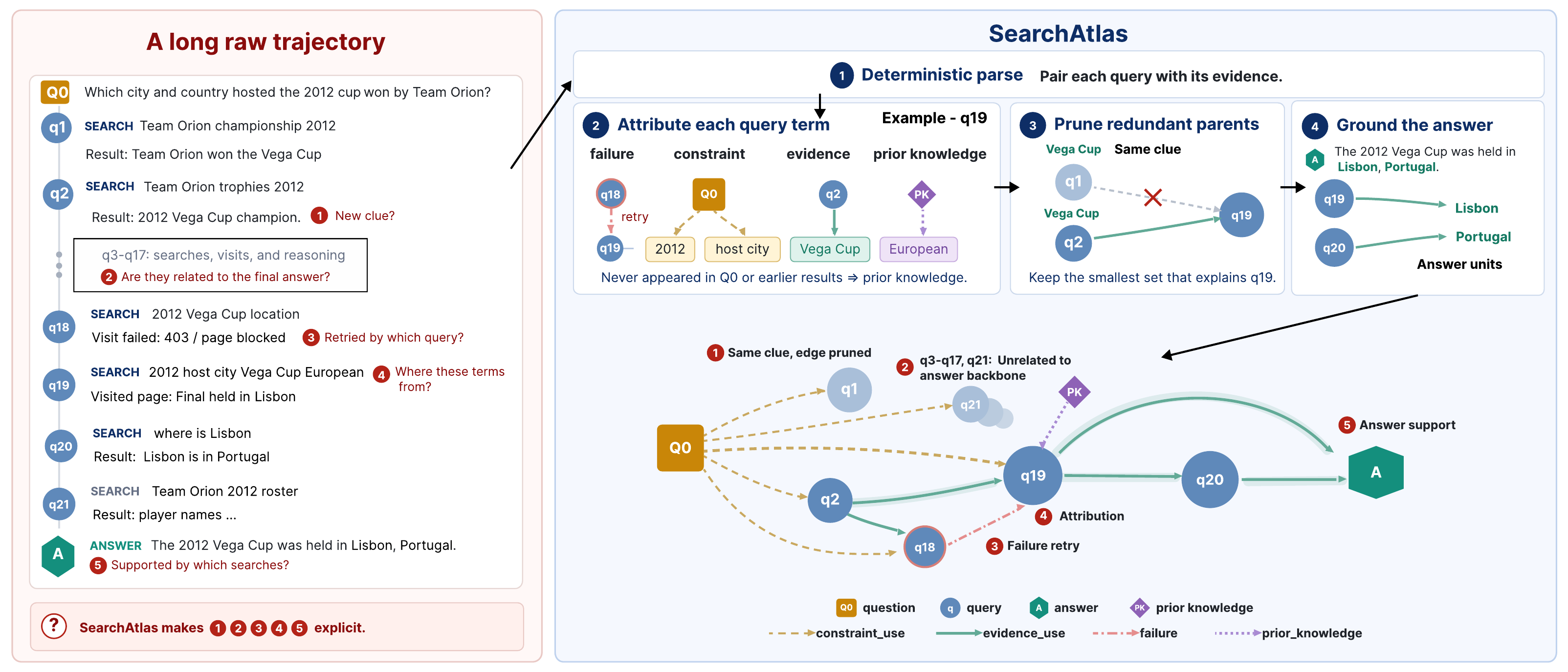}
  \caption{SearchAtlas transforms a raw search trajectory into an evidential query DAG. (a) The chronological event log hides repeated clues, verbose searches unrelated to the final answer, a potentially retried failure, query terms with no visible source, and the searches that eventually support the answer. (b) \textsc{SearchAtlas} deterministically parses the log, attributes each part of a new query to an earlier source, prunes redundant parents, and grounds answer facts. The resulting DAG keeps every query while exposing evidence flow to the final answer and turns each flagged problem into explicit typed edges.
  }
  \label{fig:searchatlas-overview}
\end{figure*}

We introduce \textsc{SearchAtlas} (\autoref{fig:searchatlas-overview}), a framework that converts each trajectory to an evidence-dependency query-to-query DAG. Its nodes represent each issued search query. Edges are added when one prior query's retrieval observably supports the formation of a later query.
We show these graphs can be automatically constructed through deterministic preprocessing that parses each query's search results and page visits into ordered reasoning nodes, followed by LLM-based attribution of the earlier evidence supporting subsequent queries and answer facts.
Against 100 human-annotated graphs, this pipeline achieves stable reconstructions and a macro edge $F_1$ ranging from $0.814$ to $0.860$ across four LLMs.

By aggregating query nodes and evidence edges into graph-level statistics, \textsc{SearchAtlas} provides a compact structural profile of an agent's strategy. Across 1,350 trajectories from five agent configurations on three benchmarks, these profiles reveal systematic differences in search scale and evidence aggregation.
For example, the median number of nodes and edges ranges from 6/7 for TYDP-Qwen3 to 103/165 for MiroThinker on BrowseComp, reflecting a significantly larger search scale of MiroThinker.
Depth, branching, and the frequency with which queries synthesize multiple earlier results also vary across settings. MiroThinker has a median DAG depth of 26, compared with 2 for TYDP-Qwen3; the multi-source query rate ranges from 8\% for TYDP-Qwen3 to 39\% for WebSailor and TYDP.

Furthermore, this structured graph representation helps diagnose whether an agent is effective in forming a plausible supporting-evidence structure relative to the question's unique constraint structure. For example, questions with \textit{sequential} constraints have to resolve intermediate uncertainty needed by later constraints, while \textit{parallel}-constraint questions can check constraints independently. The former therefore requires evidence flowing through a focused chain, while the latter prefers more direct evidence-to-answer support.

Specifically, we define three answer-support diagnostics. \textit{Answer-path topology} measures whether the evidence has the expected structure. \textit{Constraint grounding} captures whether the question's requirements are addressed in queries that contribute to the final answer, rather than only in abandoned queries whose results are not used. \textit{Prior-knowledge reliance} marks blindly trusting LLM parametric knowledge as answer-supporting evidence without validation from retrieval. Together, they localize misaligned supporting structures, unused constraints, and unsupported shortcuts to specific actions.

Across settings, trajectories with aligned answer-path topology, stronger constraint grounding, and lower prior-knowledge reliance are indicative of correct answers. We compute ROC-AUC separately within each agent by using the combined diagnostic score to rank that agent's correct and incorrect trajectories, yielding macro ROC-AUCs of $0.840$--$0.856$.
In fact, the diagnostics provide a stronger correctness signal than LLM judges given the raw trajectory or ordered query list. While this association arises because the diagnostics capture concrete failure patterns in how evidence is gathered and used, they remain informative about process quality when diverging from outcome. Our audit shows that high-scoring errors expose coherent answer-support structures that bind to the wrong target, while low-scoring successes reveal unnecessary over-search or reliance on unverified prior knowledge that happened to be correct.

\paragraph{Contributions.}
(1) We propose \textsc{SearchAtlas}, an evidence-dependency DAG representation and an automated construction pipeline validated against human annotations.
(2) Across five search agents and three benchmarks, we characterize differences in search scale and evidence aggregation and define three question-type-conditioned diagnostics.
(3) We show that these diagnostics localize process failures, provide a stronger correctness signal than unstructured trajectory baselines, and capture information complementary to final-answer accuracy.

\section{Related Work}

\paragraph{Search-agent benchmarks and evaluation.}
Benchmarks for search agents now span realistic web environments and long-horizon information-seeking tasks~\citep{mialon2023gaia,Zhou+2023,krishna2024frames,wei2025browsecomp,xi2025infodeepseek,chen2025xbench}. Most judge what the agent ultimately produces, such as task success, answer correctness, citation quality, or report quality, rather than how it arrived there~\citep{gou2025mind2web2,du2025deepresearchbench,li2026deepresearchbench2}. Several recent efforts examine the process more directly for different purposes. In the training setting, DeSA separates search optimization from answer generation after finding that answer-only rewards induce skipped retrieval and redundant queries, and Agent-RRM and PPR replace sparse outcome feedback with structured trajectory-level signals~\citep{wang2025desa,fan2026agentrm,xu2025hybridreward}. RE-TRAC summarizes accumulated evidence, uncertainties, and failures to steer later exploration~\citep{zhu2026retrac}, while other diagnostics score or localize where a trajectory becomes unreliable, exposing weaknesses invisible to output-level metrics~\citep{miromind2026miroeval,fan2026agentprocessbench,wang2026deepresearch,kim2025beyond}. These approaches treat the process as a training signal or a step-level quality judgment, but do not analyze how retrieved evidence flows through a run.

\paragraph{Graph representations of agent and reasoning processes.}
One line of work uses graph or DAG structures prescriptively, as scaffolds that models expand or schedules they execute. For reasoning, Tree-of-Thoughts~\citep{YaoToT+2023} and
Graph-of-Thoughts~\citep{Besta+2024} search over
self-generated thoughts, while DAG-Math~\citep{dziri2023faith,zhang2025dagmath} structures mathematical reasoning as a DAG for multi-agent execution. GPTSwarm and MacNet represent agent systems as optimizable graph or DAG topologies~\citep{zhuge2024gptswarm,qian2024macnet}, while Plan-over-Graph, Flash-Searcher, and S-DAG use graph or DAG schedules to parallelize execution or route reasoning across specialized agents~\citep{zhang2025planovergraph,qin2025flashsearcher,dong2026sdag}. 

\paragraph{Post-hoc graph recovery from execution traces.}
A second descriptive group recovers graph structure
post hoc from executed traces. ReasoningFlow~\citep{lee2025reasoningflow} parses reasoning traces of large reasoning models into semantically typed DAGs
to characterize motifs such as planning, reflection, and
backtracking; Graph of Verification~\citep{fang2025gov} recovers
a formal DAG from a chain-of-thought output and verifies each
node from its minimal justified premises; and
WebGraphEval~\citep{Qian+2025} is closer to our setting because it operates on web-agent trajectories. However, it aggregates actions across multiple runs into a consensus graph. \textsc{SearchAtlas} operates on one search-agent trajectory at a time. Its edges are gated by attributable retrieved evidence, such as reused snippets, visited pages, or explicit failure statements. This makes the graph an attribution object over external evidence flow, rather than a representation of internal reasoning or shared navigation behavior.

\section{\textsc{SearchAtlas}}

In this section, we describe how \textsc{SearchAtlas} converts a raw search log of the agent's queries, reasoning and tool results, into a DAG that encodes the evidential structure of the final answer.

\subsection{Graph Definition}
Let a trajectory contain $N$ search queries in chronological order. We represent its observable evidence flow as a typed directed graph $G=(V,E)$, 
\begin{gather}
    V=\{q_0,\mathrm{PK},A\}\cup\mathcal{Q}, \notag\\
    \mathcal{Q}=\{q_1,\ldots,q_N\},
    \label{eq:graph-definition}
\end{gather}
where $q_0$ is the original question, $q_i$ is the $i$-th issued query together with its retrieved search results and visited pages, $\mathrm{PK}$ is a source node for unattributed prior knowledge, and $A$ is the agent's final answer.

Each edge $(u,v)\in E$ states that the content available at $u$ observably contributes to $v$. These edges can be grouped into four types:
\textit{Constraint-use} edges ($q_0\to q_i$) indicate that query $q_i$ directly targets a requirement in the original question.
\textit{Evidence-use} edges ($q_i\to v$) mark that a later query or the final answer $v$ relies on a fact retrieved by $q_i$.
When query $q_j$ responds to an explicitly failed or insufficient earlier search $q_i$, we add \textit{failure-response} edges ($q_i\to q_j$). This accounts for both hard failures, such as zero results and blocked pages, and soft failures, where the agent judges the retrieved results insufficient.
\textit{Prior-knowledge} edges ($\mathrm{PK}\to v$) mark content in a query or the answer $v$ that emerges neither from $q_0$ nor from evidence retrieved by an earlier query.
Here $v\in\{q_{i+1},\ldots,q_N,A\}$ for an edge out of $q_i$, and $i<j$ for query-to-query edges to respect the trajectory's causal order.

For query-to-query edges, we connect an earlier query only when its retrieved content helps form the later query. If several candidates provide overlapping support (i.e., the same fact), we keep the minimal set that together explains the parts of the later query supported by earlier retrievals, avoiding redundant edges.

For edges to the final answer, we adopt more fine-grained attribution. We first split $A$ into factual units, such as names, dates, numbers, acronyms, and other key spans. For each unit, we identify a minimal set of supporting queries using a procedure similar to the one above. Each retained query receives a $q_i\to A$ edge. We denote these \textit{direct answer-support} queries by
$\mathcal{Q}_{\mathrm{ans}}=\{q_i\in\mathcal{Q}:(q_i,A)\in E\}$.
More generally, we denote an evidence-use edge as \textit{answer-reaching} if it lies on a directed path to \(A\).
If an answer unit has no retrieved support, we connect $\mathrm{PK}\to A$.

\subsection{Automated Parsing and Validation}
Manually constructing a DAG for every search trajectory is tedious labor that prevents large-scale evaluations. We show accurate automated graph parsing can be implemented. Specifically, we adopt a two-stage pipeline where deterministic preprocessing handles information that can be extracted directly from the trajectory and LLM-based attribution resolves evidence dependencies that require semantic interpretation.

First, a deterministic program extracts each search query as a node and collects its search results. It also detects search failures and adds $q_0\to q_i$ constraint-use edges when a query explicitly mentions a requirement from the original question. 

We then adopt an LLM to attribute the evidential linkages between queries. Targeting one query at a time, the LLM is given the original question, the current reasoning block, and all earlier queries with their retrieved evidence to identify incoming evidence-use edges.
The detailed attribution and parent-pruning process can be found in Appendix~\ref{app:construction-details}.

The automated pipeline is evaluated against 100 human-annotated trajectory DAGs. Comparing the predicted and human-labeled edges, the parser achieves a macro edge \(F_1\) of \(0.860\), with similar performance across the three benchmarks.
Reconstruction fidelity and downstream results remain stable across four different LLMs used for the second-stage edge attribution (Appendix~\ref{app:DAG_parsing_validation}).

\section{Process Diagnostics}
\label{sec:process-diagnostics}
The constructed DAG captures fine-grained evidence dependencies, which we use to derive high-level diagnostics of how search agents model question constraints and organize evidence. Although these dependencies merely describe how evidence flows, they become diagnostically meaningful when interpreted relative to the question's constraint structure. We therefore distinguish two question constraint types and define three type-conditioned diagnostics of answer-path topology, constraint grounding, and prior-knowledge reliance.

\subsection{Question Constraint Types}
\begin{figure}[t]
  \centering

  \includegraphics[width=\linewidth]{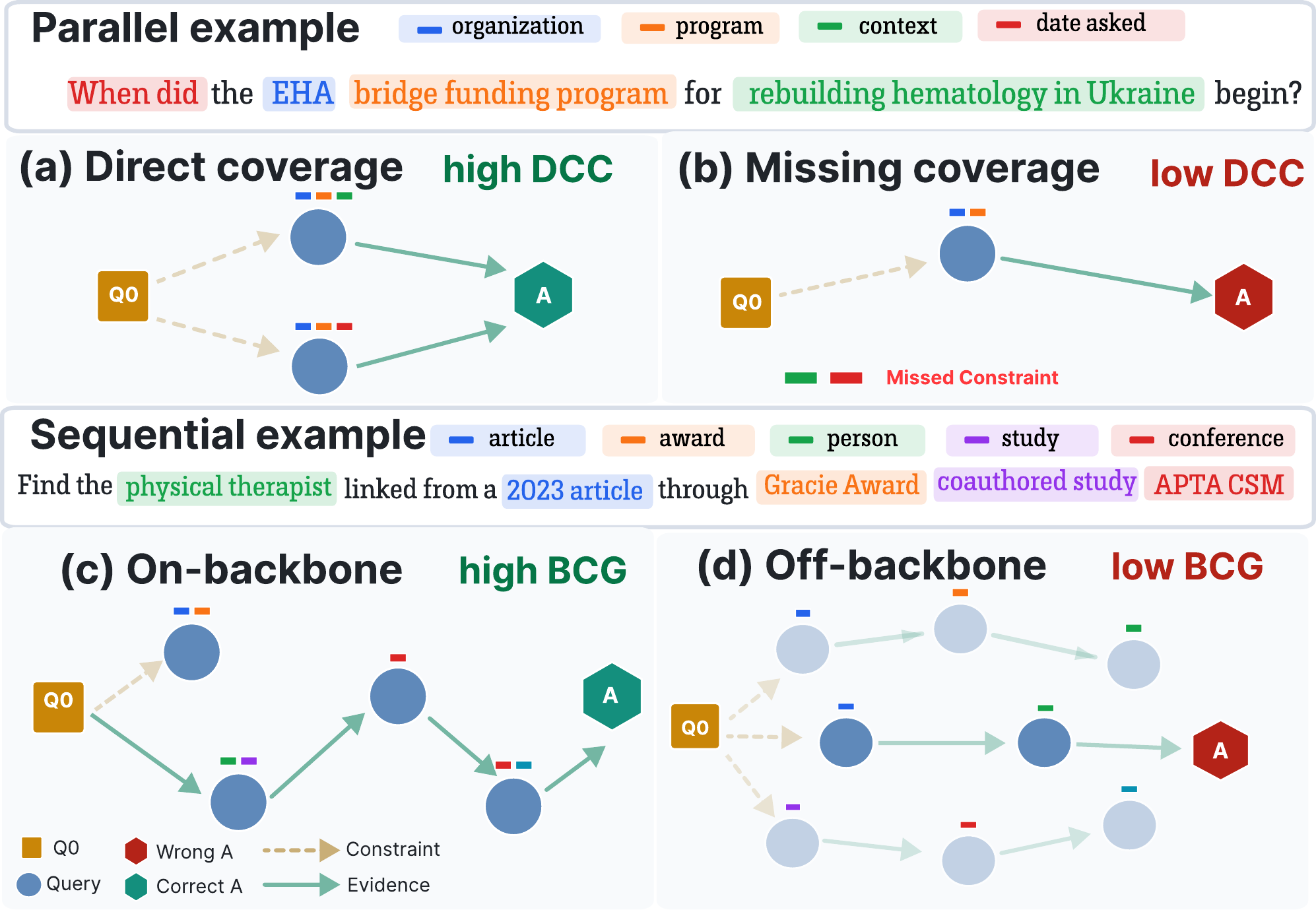}
  \caption{Illustration of constraint grounding. Constraint units are highlighted in the question examples. Four DAGs illustrate how the units are grounded on query nodes under high versus low direct constraint coverage (parallel) and high versus low backbone constraint share (sequential).}
  \label{fig:constraint-grounding-schematic}
\end{figure}

A question has \textit{sequential} constraints if they form a dependency chain in which at least one constraint refers to a latent intermediate variable. The value of this variable must be resolved before the dependent constraint can be instantiated or evaluated.
In the sequential example of \autoref{fig:constraint-grounding-schematic}, the agent needs to identify a physical therapist by tracing a 2023 article through a Gracie Award recipient to a coauthored study presented at APTA CSM. Each discovery determines the target of the next search.

Conversely, a question comes with \textit{parallel} constraints if it is specified either by one constraint or by multiple independently verifiable constraints that jointly restrict a common target variable. 
In this case, constraints may be resolved in any order.

\begin{figure}[t]
  \centering
  \includegraphics[width=\linewidth]{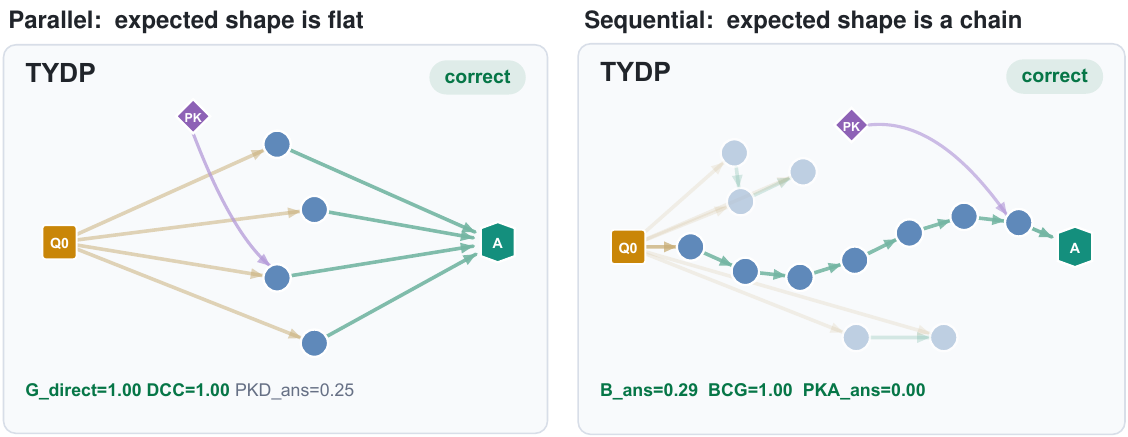}
  \caption{Ideal structures for the two question constraint types from successful TYDP runs (WebWalker-004 and BrowseComp-77). In the parallel run, every supporting query connects directly to the answer; in the sequential run, evidence concentrates in a focused chain.}
  \label{fig:reference-shapes}
\end{figure}

\subsection{Answer-Path Topology}
\label{subsec:answer-path-topology}
Motivated by prior work on supporting facts and connected reasoning paths in evidence-grounded QA evaluations~\citep{yang2018hotpotqa,trivedi2022musique,trivedi2023ircot}, we define an answer-path topology diagnostic to describe how evidence is organized to construct the final answer. We instantiate this diagnostic separately for the two question types to meet different expectations (\autoref{fig:reference-shapes}).

Since the constraints in \textit{parallel}-constraint questions can usually be checked independently, evidence for each constraint connects relatively directly to the final answer. We measure this pattern through \textbf{answer-path directness} ($G_{\mathrm{direct}}$), the fraction of answer-reaching evidence-use edges that link directly to \(A\). A higher value indicates that retrieved evidence is used directly for answer construction, rather than being routed through intermediate queries. 

For \textit{sequential}-constraint questions, effective strategies address one constraint at a time rather than wasting effort in unused evidence dispersed across side branches or abandoned paths. We measure this with \textbf{answer-backbone concentration} ($B_{\mathrm{ans}}$), the fraction of answer-reaching edges on the longest path to \(A\). High concentration indicates that most of the search effort focuses on a single chain of evidence, whereas low concentration suggests shallow exploration of constraints.

\vspace{-0.2em}

\subsection{Constraint Grounding}
\label{subsec:constraint-grounding}
We examine whether the constraints stated in the question are incorporated into the evidence paths leading to the final answer. Search agents may commit to an answer while leaving some constraints unchecked~\citep{wang2026deepresearch,ko2026illusorycompletion,lee2026kbrowsecomp}.

For \emph{sequential}-constraint questions, \textbf{backbone constraint share} (BCG) measures the fraction of constraints addressed along the longest answer-reaching path rather than dispersed across side branches. For \emph{parallel}-constraint questions, \textbf{direct constraint coverage} (DCC) measures the fraction of question constraints deployed by answer support queries $\mathcal{Q}_{\mathrm{ans}}$. High coverage indicates that the answer-path structure accounts for most of the question's identifying requirements, whereas low coverage indicates that the model may produce an answer without grounding one or more of those requirements.

\subsection{Prior-Knowledge Reliance}
\label{subsec:prior-knowledge-reliance}
Recalling facts directly from LLM parametric knowledge to construct the final answer without retrieval verification is risky due to outdated training and hallucination~\citep{chen-etal-2025-real,huang2025reinforced,lin2025adasearch}.

For \emph{sequential}-constraint questions, $\mathrm{PKA}_{\mathrm{ans}}$ is a binary indicator that monitors if prior knowledge is used directly in the final answer construction. For \emph{parallel} constraints that require independent verification, unsupported prior knowledge in any of these queries could be consequential. So we count direct answer-support queries formulated using unsupported information and normalize by the number of direct answer-support queries $|\mathcal{Q}_{\mathrm{ans}}|$.
We name it $\mathrm{PKD}_{\mathrm{ans}}$.
Higher values indicate greater reliance on unsupported information during answer construction.

\section{Experimental Results}
\label{sec:analysis_findings}

\subsection{Experimental Setup}
\paragraph{Agents.} To ensure evaluation generalizability, we select search agents with different base models and agentic scaffolding.
We evaluate (1) \textbf{WebSailor}~\citep{li2025websailor} (v1, 32B), a model post-trained with agentic reinforcement learning (DUPO) and deployed within a ReAct~\citep{YaoReAct+2023} loop, representing smaller open-source agents trained with reinforcement learning;
(2) \textbf{MiroThinker}~\citep{miromind2025mirothinker} (v1, 30B), designed for interaction scaling and trained to sustain hundreds of tool calls within a 256K-token context window;
(3) \textbf{Tongyi DeepResearch} (TYDP)~\citep{tongyi2025deepresearch}, a 30B-parameter mixture-of-experts agent trained via agentic mid-training and GRPO-based post-training for long-horizon information seeking. To separate the effect of the underlying model from that of the harness, we additionally run the Tongyi scaffold with two alternative backbones, GPT-5 and Qwen3-32B, under the same search policy; we denote these two variants as (4) \textbf{TYDP-GPT5} and (5) \textbf{TYDP-Qwen3}. This setup yields five agents in total. 

\paragraph{Datasets.}
We evaluate both question types across three benchmarks: BrowseComp~\citep{wei2025browsecomp} provides sequential-constraint questions (150), WebWalker-Hard-English~\citep{wu-etal-2025-webwalker} provides parallel-constraint questions (70), and DeepSearchQA~\citep{gupta2026deepsearchqa} provides a stress test with 25 questions per regime. Full curation details and question IDs are in Appendix~\ref{app:question-id-list}.

\subsection{\textsc{SearchAtlas} Reveals Diverse Search Behaviors}
The recovered DAGs reveal substantial differences in how extensively agents search and in how information is carried across their search steps. In Table~\ref{tab:dag-structural-statistics}, a higher number of nodes indicates that the agent takes more search steps before reaching the answer and a higher number of edges means that more search queries are interconnected through reused evidence or question constraints. MiroThinker conducts the most extensive searches on BrowseComp and DeepSearchQA, producing the most nodes and edges, whereas TYDP produces the largest DAGs on WebWalker-Hard. TYDP-Qwen3, which receives the least search training, consistently produces the shortest trajectories across all three benchmarks.

Multi-source query rate and depth show how agents integrate information differently. TYDP has the largest synthesis rate, followed by MiroThinker, on all datasets, which indicates that they tend to formulate complex search queries based on multiple evidence sources. Depth is measured by averaging the lengths of the two longest structural paths in each DAG. MiroThinker and TYDP produce the deepest DAGs, indicating the ability of both agents to build upon long chains of evidence.

\begin{table}[htbp]
\centering
\small
\setlength{\tabcolsep}{3pt}
\begin{tabular}{@{}lrrrrr@{}}
\toprule
Search Agent
& Nodes
& Edges
& Dep.
& \shortstack{Branch}
& \shortstack{Multi-Source\\Query Rate} \\
\midrule

\multicolumn{6}{@{}l}{\textbf{BrowseComp (150 questions)}} \\
\quad WebSailor
& 27 & 49 & 6 & 5 & 39\% \\
\quad MiroThinker
& 103 & 165 & 26 & 11 & 29\% \\
\quad TYDP
& 67 & 133 & 16 & 11 & 39\% \\
\quad TYDP-GPT5
& 21 & 41 & 3 & 8 & 32\% \\
\quad TYDP-Qwen3
& 6 & 7 & 2 & 2 & 8\% \\

\addlinespace[3pt]
\multicolumn{6}{@{}l}{\textbf{WebWalker-Hard (70 questions)}} \\
\quad WebSailor
& 8 & 10 & 2 & 1 & 5\% \\
\quad MiroThinker
& 8 & 12 & 3 & 2 & 6\% \\
\quad TYDP
& 16 & 23 & 3 & 4 & 11\% \\
\quad TYDP-GPT5
& 8 & 13 & 2 & 3 & 8\% \\
\quad TYDP-Qwen3
& 5 & 5 & 2 & 2 & 0\% \\

\addlinespace[3pt]
\multicolumn{6}{@{}l}{\textbf{DeepSearchQA (50 questions)}} \\
\quad WebSailor
& 13 & 19 & 3 & 3 & 15\% \\
\quad MiroThinker
& 76 & 150 & 24 & 12 & 38\% \\
\quad TYDP
& 62 & 144 & 18 & 12 & 41\% \\
\quad TYDP-GPT5
& 15 & 27 & 3 & 6 & 19\% \\
\quad TYDP-Qwen3
& 7 & 8 & 2 & 2 & 8\% \\

\bottomrule
\end{tabular}
\caption{Graph statistics of search agents across datasets.
Nodes, edges, depth, and branching are trajectory-level
medians. Branching is the average unique-child count of the
two widest non-\texttt{Q0} nodes in each trajectory, excluding
failure-derived edges. Multi-source query rate is the mean percentage
of query nodes per trajectory that rely on multiple earlier queries.
Nodes, edges, and depth are rounded to integers; branching is reported
to at most two decimal places.}
\label{tab:dag-structural-statistics}
\end{table}

\subsection{Failure Modes and Strengths of Search Strategies}
\label{subsec:metric-findings}
Our proposed diagnostics reveal the strengths and failures in how search agents construct answers.

\begin{figure*}[t]
  \centering
  \includegraphics[width=0.98\textwidth]{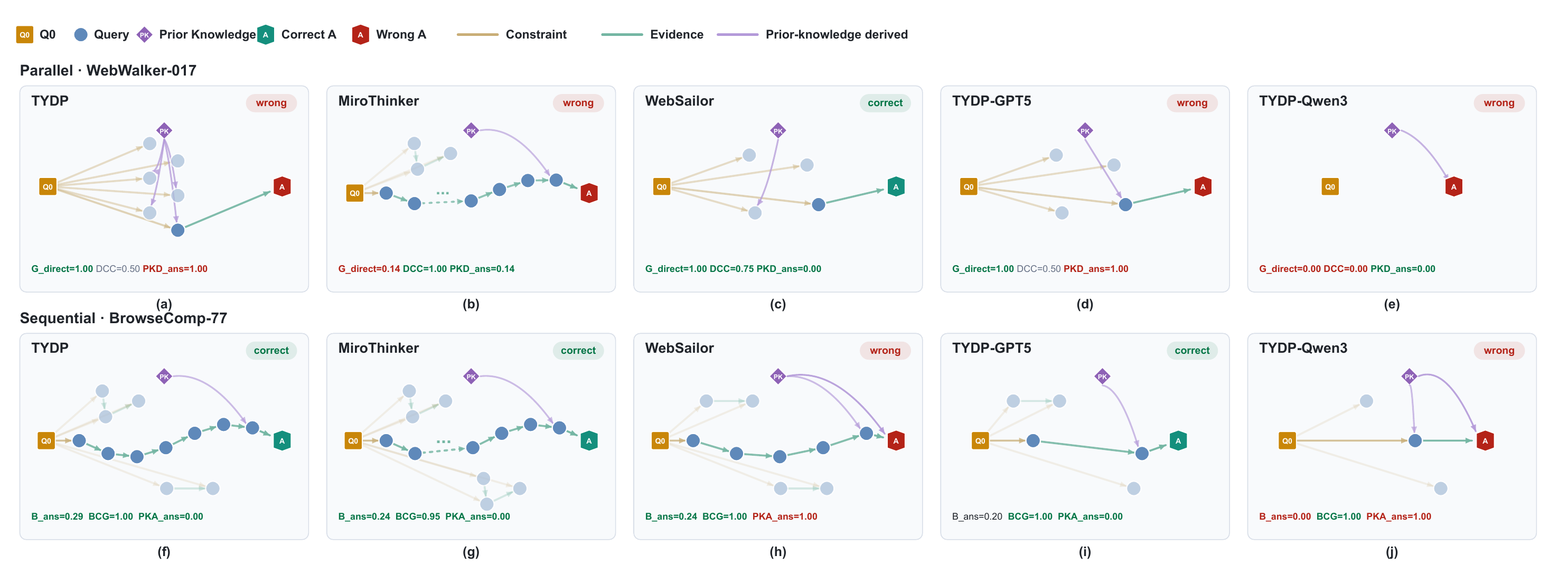}
  \caption{\textsc{SearchAtlas} visualizes trajectories of five agents on a parallel question (WebWalker-017, a--e) and a sequential question (BrowseComp-77, f--j). Query nodes shown in darker shades contribute to the final answer. The values of the three diagnostics defined in \autoref{sec:process-diagnostics} are shown below each panel; green indicates favorable values, whereas red color flags values indicating a potential process failure. The correct runs (c), (f), (g), (i) match the expected shape per question constraint type. The failed ones scatter evidence, leave constraints uncovered, or draw the answer from prior knowledge (illustrated as purple edges pointing to A).}
  \label{fig:metric-case-study}
\end{figure*}

Answer-path topology shows whether the evidence is organized the way the question requires: flat for parallel questions, a focused chain for sequential ones.
Question type induces systematic differences in DAG structure. Sequential-constraint questions tend to produce DAGs with greater depth and more branches, which lower query efficiency (the proportion of issued queries ultimately contributing to the final answer) and increase the number of queries that synthesize evidence from multiple prior steps. Parallel-constraint questions, by contrast, produce search structures in which retrieved evidence remains useful across a larger fraction of the trajectory. At the agent level, depth serves as a relatively stable signature of an agent's search behavior, while width is more sensitive to question type. These statistics reflect a fundamental difference in how evidence must be assembled to resolve each question type (Appendix~\ref{subsec:structural-profiles}), which motivates the question-type-conditioned metrics.

Parallel-constraint questions decompose into independent checks against a single candidate, so their correct evidence graphs are flat: most edges land directly on the answer node. In sequential-constraint questions, some constraints become interpretable only after earlier constraints resolve an intermediate entity. A focused dependency chain matches this resolution structure, while evidence spread over side branches, failed candidates, or abandoned directions signals fragmented support; \autoref{subsec:accuracy-metrics} tests this in aggregate.
The two question types therefore reward high answer-path directness for parallel questions and high answer-backbone concentration for sequential questions.

This contrast appears directly in the reference runs of \autoref{fig:reference-shapes}. Being flat is a matter of how evidence reaches the answer, not of how many queries supply it: the correct parallel run in panel~(c) of \autoref{fig:metric-case-study} supports the answer through a single query that carries most of the question's constraints, with no intermediate chain. Panels~(g) and~(i) share the sequential backbone shape, while the failures in panels~(e) and~(j) do not form the structure expected for their question type.

Constraint grounding shows which of the question's requirements reach the answer's support, and which are stranded in side searches or never used at all.
Topology alone is insufficient. In \autoref{fig:metric-case-study}, panels~(c) and~(d) are both flat, but panel~(d) covers only half of the constraints and fails. Panels~(h) and~(j) show the converse: constraint placement can look reasonable even when the main evidence path is missing or the answer uses unsupported information.

Prior-knowledge reliance pinpoints where unsupported content from LLM parametric knowledge enters the answer's support.
In \autoref{fig:metric-case-study}, panels~(a), (d), (h), and~(j) use unsupported information for the answer when other indicators of the graph look healthy, which potentially leads to their failures. The correct cases in the same rows do not.

\begin{table*}[t]
\centering
\small
\textit{Macro AUC after each diagnostic is added}\\[3pt]
\setlength{\tabcolsep}{7pt}
\begin{tabular}{lccccc}
\toprule
Dataset / regime & Topology only & + Grounding & Grounding gain & Full score & PK gain \\
\midrule
BrowseComp / sequential   & 0.714 & 0.816 & +0.102 & 0.855 & +0.039 \\
WebWalker-Hard / parallel & 0.746 & 0.779 & +0.033 & 0.840 & +0.061 \\
DeepSearchQA / sequential & 0.734 & 0.840 & +0.106 & 0.844 & +0.004 \\
DeepSearchQA / parallel   & 0.771 & 0.830 & +0.059 & 0.856 & +0.026 \\
\bottomrule
\end{tabular}

\vspace{0.55em}
\textit{Full-score AUC by agent}\\[3pt]
\setlength{\tabcolsep}{5.5pt}
\begin{tabular}{lrrrrrr}
\toprule
Dataset / regime & TYDP & MiroThinker & WebSailor & TYDP-GPT5 & TYDP-Qwen3 & Macro \\
\midrule
BrowseComp / sequential   & 0.898 & 0.897 & 0.878 & 0.720 & 0.880 & 0.855 \\
WebWalker-Hard / parallel & 0.822 & 0.774 & 0.884 & 0.826 & 0.895 & 0.840 \\
DeepSearchQA / sequential & 0.827 & 0.724 & 0.860 & 0.986 & 0.824 & 0.844 \\
DeepSearchQA / parallel   & 0.858 & 0.763 & 0.964 & 0.802 & 0.894 & 0.856 \\
\bottomrule
\end{tabular}
\caption{Macro AUC after adding each diagnostic and full-score AUC for each agent. Every AUC is computed within one agent; the Macro column averages the five agent-level results. Complete agent-level AUCs for all three stages are reported in Appendix~\ref{app:metric-ablation}.}
\label{tab:stagewise-auc-main}
\end{table*}

\subsection{Process Signals Reflect Answer Correctness}
\label{subsec:accuracy-metrics}

Diagnostic scores are associated with answer correctness, and the behaviors they capture are consequential for search outcomes.

\paragraph{Three diagnostics together are associated with answer correctness within each agent.}
The three diagnostics are on different scales: the topology and grounding metrics are in $[0,1]$, the sequential PK indicator is binary, and the parallel PK score can exceed $1$. We therefore normalize each metric into standard distributions and sum them (the PK reliance metric was negated) so that a higher score always means a better-supported answer. For each agent on each benchmark and regime, we compute the AUC of this score for separating the agent's correct trajectories from its incorrect ones under question-held-out evaluation; macro values average the five agent-level AUCs. TYDP-Qwen3 on BrowseComp has an answer accuracy of only $0.040$, yet its correct runs rank above its incorrect ones with an AUC of $0.880$. Topology alone already provides a strong ranking signal, and adding grounding and then PK reliance improves macro AUC on every benchmark, reaching $0.840$--$0.856$ for the full score (\autoref{tab:stagewise-auc-main}).

Furthermore, we find that a simple held-out threshold on the score predicts correctness with an $F_1$ of $0.705$ on the sequential regime and $0.780$ on the parallel regime. Calibration details and module ablations are in Appendices~\ref{app:score-calibration} and~\ref{app:metric-ablation}.
\paragraph{Comparison with non-DAG baselines.}

To confirm that \textsc{SearchAtlas} enables nontrivial, targeted diagnostics of search agent failures, we compare the DAG-derived scorer with LLM judges with access to the complete raw log or the ordered query list in context. While \textsc{SearchAtlas} simply adopts a thresholded sum of three interpretable scores, it remains the most accurate of the three (Table~\ref{tab:representation-ablation}). The full-trajectory judge predicts failures at a much higher rate and recovers fewer than half of the correct trajectories (pos. recall $0.475$ vs. $0.750$, Appendix~\ref{app:llm-judge-baseline}).
This result demonstrates that evidence dependencies provide outcome-associated information beyond unstructured trajectory content and queries alone.

\begin{table}[t]
\centering
\small
\resizebox{\linewidth}{!}{%
\begin{tabular}{lccc}
\toprule
Method & Accuracy & Macro-$F_1$ & Positive $F_1$ \\
\midrule
Ordered-query-list judge & 0.677 & 0.608 & 0.444 \\
Full-trajectory judge    & 0.738 & 0.705 & 0.606 \\
SearchAtlas              & \textbf{0.776} & \textbf{0.772} & \textbf{0.740} \\
\bottomrule
\end{tabular}}
\caption{Association with answer correctness, pooled over 1,350 trajectories.
The two non-DAG baselines are GPT-5.2 judges.}
\label{tab:representation-ablation}
\end{table}

To further test whether this signal can be explained by how much search activity a trajectory contains, we compare the DAG-derived score with a classifier using only raw-log statistics, including query count, page visits, trace length, and query redundancy (Appendix~\ref{app:raw-log-baseline}). \textsc{SearchAtlas} outperforms this baseline by a wide margin, showing that search statistics cannot substitute for modeling how evidence is organized and used.

\paragraph{Ablation of constraint-grounding region.}
To show that constraint-grounding is most important when it occurs on the answer-supporting backbone, we compare the constraint-grounding metric with a trajectory-wide baseline. This baseline marks a constraint as covered whenever any query uses it, even if that query lies in branches that do not contribute to the final answer. The constraint-grounding metric outperforms the baseline on both question types (Table~\ref{tab:constraint-localization}); on BrowseComp, the trajectory-wide baseline is no better than chance. Whether a constraint is mentioned somewhere carries little signal; what matters is whether it contributes to the answer. Appendix~\ref{app:constraint-localization-baseline} provides the complete experimental setup.

\begin{table}[t]
\centering
\small
\resizebox{\linewidth}{!}{%
\begin{tabular}{lcc}
\toprule
Dataset / regime &
\shortstack{Trajectory-wide\\coverage} &
\shortstack{DAG-localized\\grounding} \\
\midrule
BrowseComp / sequential
    & 0.459 / 0.305
    & \textbf{0.764 / 0.503} \\
WebWalker-Hard / parallel
    & 0.568 / 0.771
    & \textbf{0.680 / 0.827} \\
\bottomrule
\end{tabular}}
\caption{Constraint-grounding localization. Each cell reports macro-AUC/AP of the grounding signal alone; both columns use the same constraint units and differ only in where a deployment counts.}
\label{tab:constraint-localization}
\end{table}

\paragraph{Ablation of location of prior-knowledge use.}
We test whether the location of prior-knowledge use matters while holding all other metrics fixed. The baseline uses the same DAG, answer-path topology and constraint-grounding metrics as \textsc{SearchAtlas}. The only difference is that the baseline counts PK use anywhere in the trajectory, whereas \textsc{SearchAtlas} restricts it to the parts that contribute to the final answer. Replacing the PK score with this baseline reduces macro AUC from $0.840$ to $0.777$ on WebWalker-Hard and from $0.856$ to $0.769$ on parallel DeepSearchQA. This shows that PK is most informative when its location relative to answer support is retained (Appendix~\ref{app:pk-localization-baseline}).

\subsection{Where Process Signals and Accuracy Diverge}
\label{subsec:metric-boundary}
\begin{figure}[tb]
  \centering
  \includegraphics[width=\linewidth]{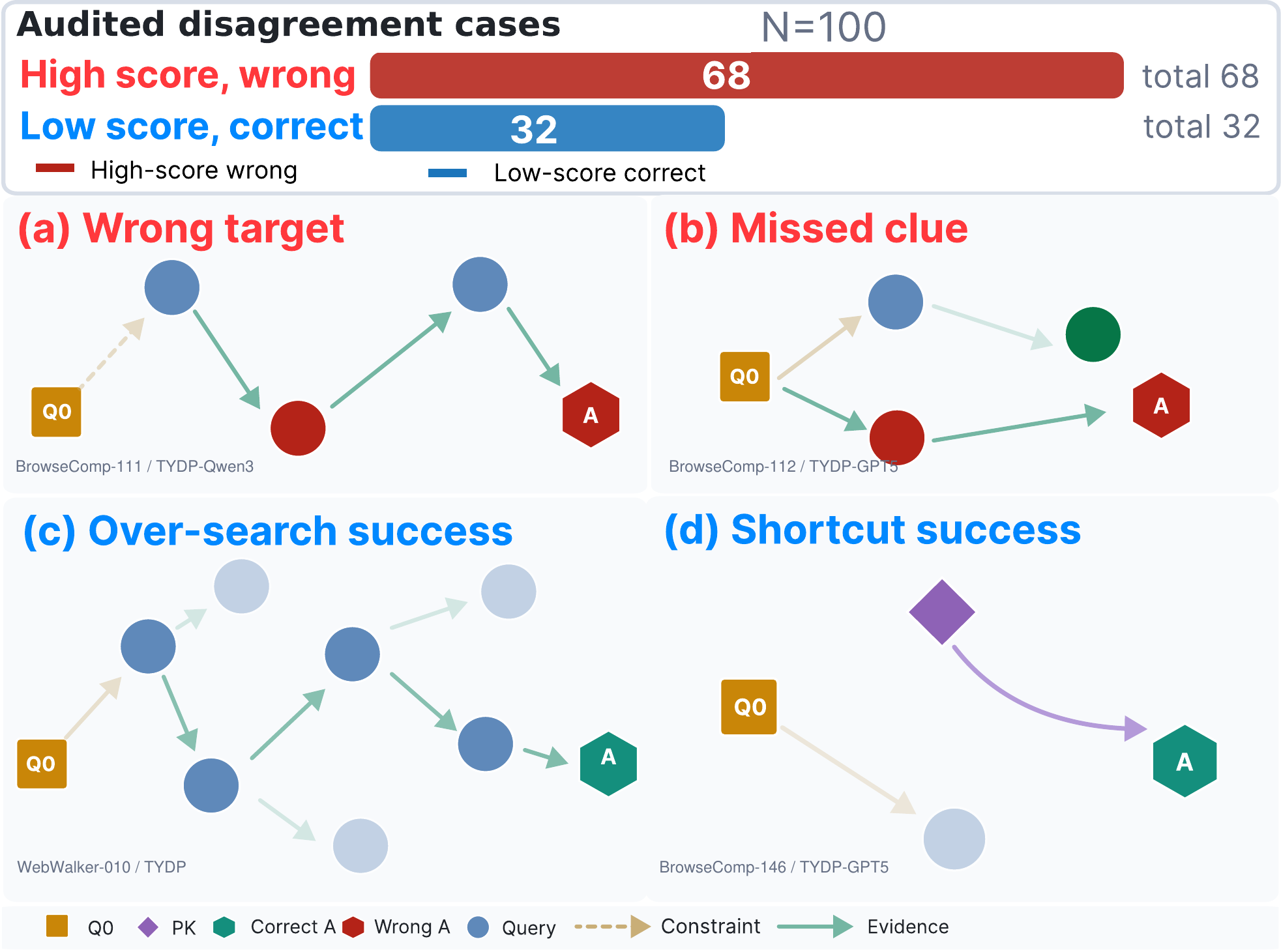}
  \caption{The top block gives the cleaned outlier distribution by score--accuracy disagreement type ($N=100$). Panels~(a)--(d) show premature target binding with unresolved contradictions, missed correct evidence, over-search success, and shortcut success.}
  \label{fig:outlier-mechanisms}
\end{figure}

Behavioral evidence structure and answer correctness can diverge in meaningful ways. A well-organized support path can be fruitless if it binds to the wrong target, while a correct answer can emerge from diffuse search or an unsupported shortcut. We audit 100 such disagreements—high-scoring incorrect and low-scoring correct trajectories—selected separately within each benchmark-regime and model (\autoref{tab:outlier-model-distribution} and \autoref{fig:outlier-mechanisms}).
\begin{table}[t]
\centering
\small
\setlength{\tabcolsep}{3.2pt}
\begin{tabular}{@{}lrrrrr@{}}
\toprule
& & \multicolumn{4}{c}{Divergence cases} \\
\cmidrule(l){3-6}
Model & \shortstack{Accuracy\\(\%)} & \shortstack{High\\wrong} & \shortstack{Low\\correct} & Total & \shortstack{Rate\\(\%)} \\
\midrule
WebSailor   & 28.9 & 18 & 3  & 21 & 7.8  \\
MiroThinker & 50.0 & 3  & 10 & 13 & 4.8  \\
TYDP        & 57.4 & 4  & 9  & 13 & 4.8  \\
TYDP-GPT5   & 58.1 & 11 & 9  & 20 & 7.4  \\
TYDP-Qwen3  & 17.8 & 32 & 1  & 33 & 12.2 \\
\bottomrule
\end{tabular}
\caption{Behavior--outcome divergence cases by agent. Rate is the percentage of all trajectories for that model.}
\label{tab:outlier-model-distribution}
\end{table}

High-scoring errors illustrate that well-organized reasoning is necessary but not sufficient for trajectory success. Our diagnostics measure whether evidence and question constraints are concentrated along the answer-reaching path, capturing the structural coherence of the trajectory. In BrowseComp-111 / TYDP-Qwen3 (\autoref{fig:outlier-mechanisms}-a), the agent links the clues to Bruno Mars and Silk Sonic without supporting evidence, then continues to gather candidate-specific evidence despite retrieving facts that contradict the question's constraints. The trajectory nevertheless scores highly because these constraints are deployed coherently along the answer backbone. This case shows that structural organization is an important component of successful reasoning but must be paired with candidate-level constraint satisfaction.

Conversely, over-search can lower the DAG-derived score despite a correct answer. WebWalker-010 / TYDP (\autoref{fig:outlier-mechanisms}-c), for example, reaches the answer through near-duplicate queries that produce diffuse support. A smaller group succeeds through prior-knowledge shortcuts rather than retrieval-backed evidence.

\section{Conclusion}
Evaluating search agents requires more than checking whether they end with the correct answer. We introduced \textsc{SearchAtlas}, a query-to-query DAG framework for post-hoc behavioral auditing of how retrieved evidence, failures, and prior knowledge appear in search trajectories. Across five search agents and three benchmarks, the recovered DAGs localize recurring process failures---fragmented answer support, question constraints that never reach the answer, and unsupported prior-knowledge shortcuts---to specific queries and edges. That these signals track answer correctness is external evidence that the failures they surface matter, and the behavior--outcome divergences show that process quality and answer correctness carry complementary information. Looking forward, the representation can be extended to analyze more general-purpose agents with richer tool settings beyond search.

\section*{Limitations}
\textsc{SearchAtlas} uses an LLM to make its local attribution decisions. Reconstruction fidelity and the downstream conclusions remain stable across four attribution models, but the recovered DAGs can still inherit systematic errors shared across models, and constructing each DAG carries a nontrivial inference cost. Scaling the framework will therefore call for cheaper attribution models, uncertainty estimates, or human-in-the-loop validation. The pipeline recovers only what the trajectory reveals. It works well when a trajectory exposes its reasoning text, tool calls, retrieved results, page visits, and final answer, but any dependency that shapes the model internally without surfacing in the log stays invisible to it. Additionally, scoring assumes the question has been routed to the right regime. Since sequential and parallel questions are judged against different expected evidence structures, a misclassification sends a question to the wrong metric family and adds evaluation noise.

Finally, our study stays within closed-answer, English-language deep-search tasks, whose questions have identifiable answers that map cleanly to sequential- or parallel-constraint regimes. This leaves open-ended, non-English and multimodal settings untested, as trajectories in these settings may have no single gold answer or fixed support structure to score against.

\section*{Acknowledgments}
We thank members of the NLP group at Duke University for fruitful discussions.
This work was supported by NSF award IIS-2211526.
\bibliography{custom}

\begin{thebibliography}{45}
\providecommand{\natexlab}[1]{#1}

\bibitem[{Besta et~al.(2024)Besta, Blach, Kubicek, Gerstenberger, Podstawski,
  Gianinazzi, Gajda, Lehmann, Niewiadomski, Nyczyk, and Hoefler}]{Besta+2024}
Maciej Besta, Nils Blach, Ales Kubicek, Robert Gerstenberger, Michal
  Podstawski, Lukas Gianinazzi, Joanna Gajda, Tomasz Lehmann, Hubert
  Niewiadomski, Piotr Nyczyk, and Torsten Hoefler. 2024.
\newblock \href {https://doi.org/10.1609/AAAI.V38I16.29720} {Graph of thoughts:
  Solving elaborate problems with large language models}.
\newblock In \emph{Thirty-Eighth {AAAI} Conference on Artificial Intelligence,
  {AAAI} 2024, Thirty-Sixth Conference on Innovative Applications of Artificial
  Intelligence, {IAAI} 2024, Fourteenth Symposium on Educational Advances in
  Artificial Intelligence, {EAAI} 2024, February 20-27, 2024, Vancouver,
  Canada}, pages 17682--17690. {AAAI} Press.

\bibitem[{Chen et~al.(2025{\natexlab{a}})Chen, Ren, Liu, Hu, Tian, Xie, Liu,
  Zhang, Liu, Gong, Sun, Hou, Yang, Pan, Lou, Mao, Liu, Li, Liu, Liu, Wang, Li,
  Niu, Zhang, Yan, Wang, Zhang, Hung, Jiang, Liu, Yin, Ma, and
  Mo}]{chen2025xbench}
Kaiyuan Chen, Yixin Ren, Yang Liu, Xiaobo Hu, Haotong Tian, Tianbao Xie, Fangfu
  Liu, Haoye Zhang, Hongzhang Liu, Yuan Gong, Chen Sun, Han Hou, Hui Yang,
  James Pan, Jianan Lou, Jiayi Mao, Jizheng Liu, Jinpeng Li, Kangyi Liu, and 14
  others. 2025{\natexlab{a}}.
\newblock \href {https://arxiv.org/abs/2506.13651} {xbench: Tracking agents
  productivity scaling with profession-aligned real-world evaluations}.
\newblock \emph{Preprint}, arXiv:2506.13651.

\bibitem[{Chen et~al.(2025{\natexlab{b}})Chen, Huang, and
  Dhingra}]{chen-etal-2025-real}
Sanxing Chen, Yukun Huang, and Bhuwan Dhingra. 2025{\natexlab{b}}.
\newblock \href {https://doi.org/10.18653/v1/2025.acl-long.81} {Real-time
  factuality assessment from adversarial feedback}.
\newblock In \emph{Proceedings of the 63rd Annual Meeting of the Association
  for Computational Linguistics (Volume 1: Long Papers)}, pages 1610--1630,
  Vienna, Austria. Association for Computational Linguistics.

\bibitem[{Dong et~al.(2026)Dong, Lin, Lin, and Zhang}]{dong2026sdag}
Jiangwen Dong, Zehui Lin, Wanyu Lin, and Mingjin Zhang. 2026.
\newblock \href {https://doi.org/10.1609/aaai.v40i35.40180} {{S-DAG}: A
  subject-based directed acyclic graph for multi-agent heterogeneous
  reasoning}.
\newblock \emph{Proceedings of the AAAI Conference on Artificial Intelligence},
  40(35):29394--29402.

\bibitem[{Du et~al.(2025)Du, Xu, Zhu, Wang, and Mao}]{du2025deepresearchbench}
Mingxuan Du, Benfeng Xu, Chiwei Zhu, Xiaorui Wang, and Zhendong Mao. 2025.
\newblock \href {https://arxiv.org/abs/2506.11763} {Deepresearch bench: A
  comprehensive benchmark for deep research agents}.
\newblock \emph{Preprint}, arXiv:2506.11763.

\bibitem[{Dziri et~al.(2023)Dziri, Lu, Sclar, Li, Jiang, Lin, West,
  Bhagavatula, Le~Bras, Hwang, Sanyal, Welleck, Ren, Ettinger, Harchaoui, and
  Choi}]{dziri2023faith}
Nouha Dziri, Ximing Lu, Melanie Sclar, Xiang~Lorraine Li, Liwei Jiang,
  Bill~Yuchen Lin, Peter West, Chandra Bhagavatula, Ronan Le~Bras, Jena~D.
  Hwang, Soumya Sanyal, Sean Welleck, Xiang Ren, Allyson Ettinger, Zaid
  Harchaoui, and Yejin Choi. 2023.
\newblock \href {https://arxiv.org/abs/2305.18654} {Faith and fate: Limits of
  transformers on compositionality}.
\newblock In \emph{Advances in Neural Information Processing Systems
  (NeurIPS)}.

\bibitem[{Fan et~al.(2026{\natexlab{a}})Fan, Feng, Zhang, Peng, Li, Jiang,
  Chen, Pei, Cai, and Yue}]{fan2026agentrm}
Kaixuan Fan, Kaituo Feng, Manyuan Zhang, Tianshuo Peng, Zhixun Li, Yilei Jiang,
  Shuang Chen, Peng Pei, Xunliang Cai, and Xiangyu Yue. 2026{\natexlab{a}}.
\newblock \href {https://arxiv.org/abs/2601.22154} {Exploring reasoning reward
  model for agents}.
\newblock \emph{Preprint}, arXiv:2601.22154.

\bibitem[{Fan et~al.(2026{\natexlab{b}})Fan, Ye, Huo, Chen, Guo, Yang, Yang,
  Ye, Chen, Chen, Cong, and Lin}]{fan2026agentprocessbench}
Shengda Fan, Xuyan Ye, Yupeng Huo, Zhi-Yuan Chen, Yiju Guo, Shenzhi Yang,
  Wenkai Yang, Shuqi Ye, Jingwen Chen, Haotian Chen, Xin Cong, and Yankai Lin.
  2026{\natexlab{b}}.
\newblock {AgentProcessBench}: Diagnosing step-level process quality in
  tool-using agents.
\newblock \emph{arXiv preprint arXiv:2603.14465}.

\bibitem[{Fang et~al.(2026)Fang, Zhang, Wang, Wan, and Xu}]{fang2025gov}
Jiwei Fang, Bin Zhang, Changwei Wang, Jin Wan, and Zhiwei Xu. 2026.
\newblock \href {https://doi.org/10.1609/aaai.v40i36.40322} {Graph of
  verification: Structured verification of llm reasoning with directed acyclic
  graphs}.
\newblock In \emph{Proceedings of the AAAI Conference on Artificial
  Intelligence}, volume~40, pages 30665--30672.

\bibitem[{Gou et~al.(2025)Gou, Huang, Ning, Gu, Lin, Qi, Kopanev, Yu,
  Gutiérrez, Shu, Song, Wu, Chen, Moussa, Zhang, Xie, Li, Xue, Liao, Zhang,
  Zheng, Cai, Rozgic, Ziyadi, Sun, and Su}]{gou2025mind2web2}
Boyu Gou, Zanming Huang, Yuting Ning, Yu~Gu, Michael Lin, Weijian Qi, Andrei
  Kopanev, Botao Yu, Bernal~Jiménez Gutiérrez, Yiheng Shu, Chan~Hee Song,
  Jiaman Wu, Shijie Chen, Hanane~Nour Moussa, Tianshu Zhang, Jian Xie, Yifei
  Li, Tianci Xue, Zeyi Liao, and 7 others. 2025.
\newblock \href {https://arxiv.org/abs/2506.21506} {Mind2web 2: Evaluating
  agentic search with agent-as-a-judge}.

\bibitem[{Guo and Vosoughi(2025)}]{guo2025serial}
Xiaobo Guo and Soroush Vosoughi. 2025.
\newblock Serial position effects of large language models.
\newblock In \emph{Findings of the Association for Computational Linguistics:
  ACL 2025}, pages 927--953, Vienna, Austria.

\bibitem[{Gupta et~al.(2026)Gupta, Chatterjee, Haas, Tao, Wang, Liu, Oiwa,
  Gribovskaya, Ackermann, Blitzer, Goldshtein, and Das}]{gupta2026deepsearchqa}
Nikita Gupta, Riju Chatterjee, Lukas Haas, Connie Tao, Andrew Wang, Chang Liu,
  Hidekazu Oiwa, Elena Gribovskaya, Jan Ackermann, John Blitzer, Sasha
  Goldshtein, and Dipanjan Das. 2026.
\newblock \href {https://arxiv.org/abs/2601.20975} {Deepsearchqa: Bridging the
  comprehensiveness gap for deep research agents}.

\bibitem[{Huang et~al.(2025)Huang, Yuan, Ju, Zhao, and
  Liu}]{huang2025reinforced}
Ziyang Huang, Xiaowei Yuan, Yiming Ju, Jun Zhao, and Kang Liu. 2025.
\newblock \href {https://doi.org/10.48550/arXiv.2505.07596} {Reinforced
  internal-external knowledge synergistic reasoning for efficient adaptive
  search agent}.
\newblock \emph{Preprint}, arXiv:2505.07596.

\bibitem[{Kim et~al.(2025)Kim, Park, In, Kim, Lee, and Park}]{kim2025beyond}
Wonjoong Kim, Sang~Yoon Park, Yeonjun In, Sein Kim, Dongha Lee, and Chanyoung
  Park. 2025.
\newblock Beyond the final answer: Evaluating the reasoning trajectories of
  tool-augmented agents.
\newblock \emph{arXiv preprint arXiv:2510.02837}.

\bibitem[{Ko et~al.(2026)Ko, Kim, Kim, Park, Lee, Kim, Lee, and
  Lee}]{ko2026illusorycompletion}
Dayoon Ko, Jihyuk Kim, Sohyeon Kim, Haeju Park, Dahyun Lee, Gunhee Kim, Moontae
  Lee, and Kyungjae Lee. 2026.
\newblock \href {https://doi.org/10.48550/arXiv.2602.07549} {When is enough not
  enough? illusory completion in search agents}.
\newblock \emph{Preprint}, arXiv:2602.07549.

\bibitem[{Krishna et~al.(2024)Krishna, Krishna, Mohananey, Schwarcz, Stambler,
  Upadhyay, and Faruqui}]{krishna2024frames}
Satyapriya Krishna, Kalpesh Krishna, Anhad Mohananey, Steven Schwarcz, Adam
  Stambler, Shyam Upadhyay, and Manaal Faruqui. 2024.
\newblock Fact, fetch, and reason: A unified evaluation of retrieval-augmented
  generation.
\newblock \emph{arXiv preprint arXiv:2409.12941}.

\bibitem[{Lee et~al.(2025)Lee, Mukherjee, Hakkani-Tur, and
  Hockenmaier}]{lee2025reasoningflow}
Jinu Lee, Sagnik Mukherjee, Dilek Hakkani-Tur, and Julia Hockenmaier. 2025.
\newblock \href {https://arxiv.org/abs/2506.02532} {Reasoningflow: Semantic
  structure of complex reasoning traces}.
\newblock \emph{Preprint}, arXiv:2506.02532.

\bibitem[{Lee et~al.(2026)Lee, Yoon, Son, Kim, Ko, Park, Yoo, Cho, Park, Lee,
  Jang, Kim, Kim, Cho, and Kim}]{lee2026kbrowsecomp}
Nahyun Lee, Dongkeun Yoon, Guijin Son, Geewook Kim, Dayoon Ko, Jeonghun Park,
  Haneul Yoo, Jaewon Cho, Junghun Park, Changyoon Lee, Kyochul Jang, Jaeyeon
  Kim, Eunsu Kim, Woojin Cho, and Seungone Kim. 2026.
\newblock \href {https://doi.org/10.48550/arXiv.2606.02404} {{K-BrowseComp}: A
  web browsing agent benchmark grounded in korean contexts}.
\newblock \emph{Preprint}, arXiv:2606.02404.

\bibitem[{Li et~al.(2025)Li, Zhang, Yin, Zhang, Ou, Wu, Yin, Li, Tao, Wang,
  Shen, Zhang, Zhang, Wu, Jiang, Yan, Xie, Huang, and Zhou}]{li2025websailor}
Kuan Li, Zhongwang Zhang, Huifeng Yin, Liwen Zhang, Litu Ou, Jialong Wu,
  Wenbiao Yin, Baixuan Li, Zhengwei Tao, Xinyu Wang, Weizhou Shen, Junkai
  Zhang, Dingchu Zhang, Xixi Wu, Yong Jiang, Ming Yan, Pengjun Xie, Fei Huang,
  and Jingren Zhou. 2025.
\newblock \href {https://arxiv.org/abs/2507.02592} {Websailor: Navigating
  super-human reasoning for web agent}.
\newblock \emph{Preprint}, arXiv:2507.02592.

\bibitem[{Li et~al.(2026)Li, Du, Xu, Zhu, Wang, and
  Mao}]{li2026deepresearchbench2}
Ruizhe Li, Mingxuan Du, Benfeng Xu, Chiwei Zhu, Xiaorui Wang, and Zhendong Mao.
  2026.
\newblock \href {https://arxiv.org/abs/2601.08536} {Deepresearch bench ii:
  Diagnosing deep research agents via rubrics from expert report}.

\bibitem[{Lin et~al.(2025)Lin, Chen, Li, Lee, Chen, and
  Meng}]{lin2025adasearch}
Tzu-Han Lin, Wei-Lin Chen, Chen-An Li, Hung-yi Lee, Yun-Nung Chen, and Yu~Meng.
  2025.
\newblock \href {https://doi.org/10.48550/arXiv.2512.16883} {Adasearch:
  Balancing parametric knowledge and search in large language models via
  reinforcement learning}.
\newblock \emph{Preprint}, arXiv:2512.16883.

\bibitem[{Liu et~al.(2024)Liu, Lin, Hewitt, Paranjape, Bevilacqua, Petroni, and
  Liang}]{liu2024lost}
Nelson~F. Liu, Kevin Lin, John Hewitt, Ashwin Paranjape, Michele Bevilacqua,
  Fabio Petroni, and Percy Liang. 2024.
\newblock Lost in the middle: How language models use long contexts.
\newblock \emph{Transactions of the Association for Computational Linguistics},
  12:157--173.

\bibitem[{Mialon et~al.(2023)Mialon, Fourrier, Swift, Wolf, LeCun, and
  Scialom}]{mialon2023gaia}
Gr{\'e}goire Mialon, Cl{\'e}mentine Fourrier, Craig Swift, Thomas Wolf, Yann
  LeCun, and Thomas Scialom. 2023.
\newblock {GAIA}: A benchmark for general {AI} assistants.
\newblock \emph{arXiv preprint arXiv:2311.12983}.

\bibitem[{Qian et~al.(2024)Qian, Xie, Wang, Liu, Dang, Du, Chen, Yang, Liu, and
  Sun}]{qian2024macnet}
Chen Qian, Zihao Xie, Yifei Wang, Wei Liu, Yufan Dang, Zhuoyun Du, Weize Chen,
  Cheng Yang, Zhiyuan Liu, and Maosong Sun. 2024.
\newblock \href {https://arxiv.org/abs/2406.07155} {Scaling
  large-language-model-based multi-agent collaboration}.
\newblock \emph{Preprint}, arXiv:2406.07155.

\bibitem[{Qian et~al.(2025)Qian, Wang, Zhang, Zong, Chen, Zhou, Huang, Zeng,
  Hu, Song, and Zhang}]{Qian+2025}
Yaoyao Qian, Yuanli Wang, Jinda Zhang, Yun Zong, Meixu Chen, Hanhan Zhou,
  Jindan Huang, Yifan Zeng, Xinyu Hu, Chan~Hee Song, and Danqing Zhang. 2025.
\newblock \href {https://arxiv.org/abs/2510.19205} {Webgrapheval: Multi-turn
  trajectory evaluation for web agents using graph representation}.

\bibitem[{Qin et~al.(2025)Qin, Chen, Wang, Xing, Zhu, Zhu, Shi, Liu, Zhang,
  Liu, Jiang, Gao, and Zhou}]{qin2025flashsearcher}
Tianrui Qin, Qianben Chen, Sinuo Wang, He~Xing, King Zhu, He~Zhu, Dingfeng Shi,
  Xinxin Liu, Ge~Zhang, Jiaheng Liu, Yuchen~Eleanor Jiang, Xitong Gao, and
  Wangchunshu Zhou. 2025.
\newblock \href {https://arxiv.org/abs/2509.25301} {Flash-searcher: Fast and
  effective web agents via dag-based parallel execution}.
\newblock \emph{Preprint}, arXiv:2509.25301.

\bibitem[{Team et~al.(2025{\natexlab{a}})Team, Bai, Bing, Chen, Chen, Chen,
  Chen, Chen, Dai, Dong, Dou, Deng, Fu, Ge, Han, Huang, Huang, Jiao, Jiang,
  Jiao, Jian, Lei, Li, Luo, Li, Lin, Liu, Li, Ni, Ren, Sun, Su, Tao, Wang,
  Wang, Wang, Wang, Wang, Wang, Wang, Wang, Wang, Wang, Xu, Xing, Yang, Ye, Yu,
  Yu, Zhong, Zhao, Zhu, Zhou, Zhang, and Zhu}]{miromind2025mirothinker}
MiroMind Team, Song Bai, Lidong Bing, Carson Chen, Guanzheng Chen, Yuntao Chen,
  Zhe Chen, Ziyi Chen, Jifeng Dai, Xuan Dong, Wenhan Dou, Yue Deng, Yunjie Fu,
  Junqi Ge, Chenxia Han, Tammy Huang, Zhenhang Huang, Jerry Jiao, Shilei Jiang,
  and 36 others. 2025{\natexlab{a}}.
\newblock \href {https://arxiv.org/abs/2511.11793} {Mirothinker: Pushing the
  performance boundaries of open-source research agents via model, context, and
  interactive scaling}.
\newblock \emph{Preprint}, arXiv:2511.11793.

\bibitem[{Team et~al.(2025{\natexlab{b}})Team, Li, Zhang, Zhang, Huang, Li,
  Chen, Yin, Wu, Zhou, Li, Su, Ou, Zhang, Xie, Ye, Yin, Yu, Wang, Wu, Chen,
  Zhao, Zhang, Tao, Zhang, Qiao, Wang, Yu, Fu, Shen, Yang, Lin, Zhang, Zeng,
  Yang, Yin, Song, Yan, Liao, Xia, Xiao, Min, Ding, Fang, Chen, Huang, Wang,
  Cai, Shen, Wang, Guan, Geng, Shi, Wu, Chen, Li, and
  Jiang}]{tongyi2025deepresearch}
Tongyi~DeepResearch Team, Baixuan Li, Bo~Zhang, Dingchu Zhang, Fei Huang,
  Guangyu Li, Guoxin Chen, Huifeng Yin, Jialong Wu, Jingren Zhou, Kuan Li,
  Liangcai Su, Litu Ou, Liwen Zhang, Pengjun Xie, Rui Ye, Wenbiao Yin, Xinmiao
  Yu, Xinyu Wang, and 38 others. 2025{\natexlab{b}}.
\newblock \href {https://arxiv.org/abs/2510.24701} {Tongyi deepresearch
  technical report}.
\newblock \emph{Preprint}, arXiv:2510.24701.

\bibitem[{Trivedi et~al.(2022)Trivedi, Balasubramanian, Khot, and
  Sabharwal}]{trivedi2022musique}
Harsh Trivedi, Niranjan Balasubramanian, Tushar Khot, and Ashish Sabharwal.
  2022.
\newblock \href {https://doi.org/10.1162/tacl_a_00475} {{M}u{S}i{Q}ue: Multihop
  questions via single-hop question composition}.
\newblock \emph{Transactions of the Association for Computational Linguistics},
  10:539--554.

\bibitem[{Trivedi et~al.(2023)Trivedi, Balasubramanian, Khot, and
  Sabharwal}]{trivedi2023ircot}
Harsh Trivedi, Niranjan Balasubramanian, Tushar Khot, and Ashish Sabharwal.
  2023.
\newblock \href {https://doi.org/10.18653/v1/2023.acl-long.557} {Interleaving
  retrieval with chain-of-thought reasoning for knowledge-intensive multi-step
  questions}.
\newblock In \emph{Proceedings of the 61st Annual Meeting of the Association
  for Computational Linguistics (Volume 1: Long Papers)}, pages 10014--10037,
  Toronto, Canada. Association for Computational Linguistics.

\bibitem[{Wang et~al.(2026)Wang, Feng, Wu, Li, Xie, Ren, Zhu, Han, Meng, Feng,
  and Liu}]{wang2026deepresearch}
Jiaming Wang, Ziteng Feng, Jiangtao Wu, Ruihao Li, Qianqian Xie, Yuxiang Ren,
  He~Zhu, Xueming Han, Fanyu Meng, Junlan Feng, and Jiaheng Liu. 2026.
\newblock \href {https://doi.org/10.48550/arXiv.2606.02060} {Where do
  deep-research agents go wrong? span-level error localization in agent
  trajectories}.
\newblock \emph{Preprint}, arXiv:2606.02060.

\bibitem[{Wang et~al.(2025)Wang, Wei, Zhu, and Meng}]{wang2025desa}
Yiding Wang, Zhepei Wei, Xinyu Zhu, and Yu~Meng. 2025.
\newblock \href {https://arxiv.org/abs/2510.04695} {Beyond outcome reward:
  Decoupling search and answering improves llm agents}.
\newblock \emph{Preprint}, arXiv:2510.04695.

\bibitem[{Wei et~al.(2025)Wei, Sun, Papay, McKinney, Han, Fulford, Chung,
  Passos, Fedus, and Glaese}]{wei2025browsecomp}
Jason Wei, Zhiqing Sun, Spencer Papay, Scott McKinney, Jeffrey Han, Isa
  Fulford, Hyung~Won Chung, Alex~Tachard Passos, William Fedus, and Amelia
  Glaese. 2025.
\newblock \href {https://doi.org/10.48550/arXiv.2504.12516} {Browsecomp: A
  simple yet challenging benchmark for browsing agents}.
\newblock \emph{arXiv preprint arXiv:2504.12516}.

\bibitem[{Wu et~al.(2025)Wu, Yin, Jiang, Wang, Xi, Fang, Zhang, He, Zhou, Xie,
  and Huang}]{wu-etal-2025-webwalker}
Jialong Wu, Wenbiao Yin, Yong Jiang, Zhenglin Wang, Zekun Xi, Runnan Fang,
  Linhai Zhang, Yulan He, Deyu Zhou, Pengjun Xie, and Fei Huang. 2025.
\newblock \href {https://doi.org/10.18653/v1/2025.acl-long.508} {{W}eb{W}alker:
  Benchmarking {LLM}s in web traversal}.
\newblock In \emph{Proceedings of the 63rd Annual Meeting of the Association
  for Computational Linguistics (Volume 1: Long Papers)}, pages 10290--10305,
  Vienna, Austria. Association for Computational Linguistics.

\bibitem[{Xi et~al.(2025)Xi, Lin, Zhu, Xiao, Ou, Liu, Wan, Chen, Liu, Wang,
  Tang, Zhang, and Yu}]{xi2025infodeepseek}
Yunjia Xi, Jianghao Lin, Menghui Zhu, Yongzhao Xiao, Zhuoying Ou, Jiaqi Liu,
  Tong Wan, Bo~Chen, Weiwen Liu, Yasheng Wang, Ruiming Tang, Weinan Zhang, and
  Yong Yu. 2025.
\newblock \href {https://arxiv.org/abs/2505.15872} {Infodeepseek: Benchmarking
  agentic information seeking for retrieval-augmented generation}.
\newblock \emph{Preprint}, arXiv:2505.15872.

\bibitem[{Xu et~al.(2025)Xu, Li, Xing, Zhang, Li, and Shi}]{xu2025hybridreward}
Peiran Xu, Zhuohao Li, Xiaoying Xing, Guannan Zhang, Debiao Li, and Kunyu Shi.
  2025.
\newblock \href {https://arxiv.org/abs/2509.25598} {Hybrid reward normalization
  for process-supervised non-verifiable agentic tasks}.
\newblock \emph{Preprint}, arXiv:2509.25598.

\bibitem[{Yang et~al.(2018)Yang, Qi, Zhang, Bengio, Cohen, Salakhutdinov, and
  Manning}]{yang2018hotpotqa}
Zhilin Yang, Peng Qi, Saizheng Zhang, Yoshua Bengio, William Cohen, Ruslan
  Salakhutdinov, and Christopher~D. Manning. 2018.
\newblock \href {https://doi.org/10.18653/v1/D18-1259} {{H}otpot{QA}: A dataset
  for diverse, explainable multi-hop question answering}.
\newblock In \emph{Proceedings of the 2018 Conference on Empirical Methods in
  Natural Language Processing}, pages 2369--2380, Brussels, Belgium.
  Association for Computational Linguistics.

\bibitem[{Yao et~al.(2023{\natexlab{a}})Yao, Yu, Zhao, Shafran, Griffiths, Cao,
  and Narasimhan}]{YaoToT+2023}
Shunyu Yao, Dian Yu, Jeffrey Zhao, Izhak Shafran, Thomas~L. Griffiths, Yuan
  Cao, and Karthik Narasimhan. 2023{\natexlab{a}}.
\newblock \href
  {http://papers.nips.cc/paper\_files/paper/2023/hash/271db9922b8d1f4dd7aaef84ed5ac703-Abstract-Conference.html}
  {Tree of thoughts: Deliberate problem solving with large language models}.
\newblock In \emph{Advances in Neural Information Processing Systems 36: Annual
  Conference on Neural Information Processing Systems 2023, NeurIPS 2023, New
  Orleans, LA, USA, December 10 - 16, 2023}.

\bibitem[{Yao et~al.(2023{\natexlab{b}})Yao, Zhao, Yu, Du, Shafran, Narasimhan,
  and Cao}]{YaoReAct+2023}
Shunyu Yao, Jeffrey Zhao, Dian Yu, Nan Du, Izhak Shafran, Karthik~R.
  Narasimhan, and Yuan Cao. 2023{\natexlab{b}}.
\newblock \href {https://openreview.net/pdf?id=WE\_vluYUL-X} {React:
  Synergizing reasoning and acting in language models}.
\newblock In \emph{The Eleventh International Conference on Learning
  Representations, {ICLR} 2023, Kigali, Rwanda, May 1-5, 2023}. OpenReview.net.

\bibitem[{Ye et~al.(2026)Ye, Hu, Zhu, Li, Jin, Xiao, Wang, Wang, Zhang, Wang,
  Deng, Wang, Zhang, Su, Wang, Zhao, Wei, Ren, Hooi, Bo, Yan, and
  Bing}]{miromind2026miroeval}
Fangda Ye, Yuxin Hu, Pengxiang Zhu, Yibo Li, Ziqi Jin, Yao Xiao, Yibo Wang, Lei
  Wang, Zhen Zhang, Lu~Wang, Yue Deng, Bin Wang, Yifan Zhang, Liangcai Su,
  Xinyu Wang, He~Zhao, Chen Wei, Qiang Ren, Bryan Hooi, and 3 others. 2026.
\newblock \href {https://arxiv.org/abs/2603.28407} {Miroeval: Benchmarking
  multimodal deep research agents in process and outcome}.
\newblock \emph{Preprint}, arXiv:2603.28407.

\bibitem[{Zhang et~al.(2025{\natexlab{a}})Zhang, Ma, Cao, Zhang, and
  Zhao}]{zhang2025planovergraph}
Shiqi Zhang, Xinbei Ma, Zouying Cao, Zhuosheng Zhang, and Hai Zhao.
  2025{\natexlab{a}}.
\newblock \href {https://arxiv.org/abs/2502.14563} {Plan-over-graph: Towards
  parallelable llm agent schedule}.
\newblock \emph{Preprint}, arXiv:2502.14563.

\bibitem[{Zhang et~al.(2025{\natexlab{b}})Zhang, Kuzborskij, Lee, Leng, and
  Liu}]{zhang2025dagmath}
Yuanhe Zhang, Ilja Kuzborskij, Jason~D. Lee, Chenlei Leng, and Fanghui Liu.
  2025{\natexlab{b}}.
\newblock \href {https://arxiv.org/abs/2510.19842} {Dag-math: Graph-of-thought
  guided mathematical reasoning in llms}.
\newblock \emph{Preprint}, arXiv:2510.19842.

\bibitem[{Zhou et~al.(2024)Zhou, Xu, Zhu, Zhou, Lo, Sridhar, Cheng, Ou, Bisk,
  Fried, Alon, and Neubig}]{Zhou+2023}
Shuyan Zhou, Frank~F. Xu, Hao Zhu, Xuhui Zhou, Robert Lo, Abishek Sridhar,
  Xianyi Cheng, Tianyue Ou, Yonatan Bisk, Daniel Fried, Uri Alon, and Graham
  Neubig. 2024.
\newblock \href {https://arxiv.org/abs/2307.13854} {Webarena: A realistic web
  environment for building autonomous agents}.
\newblock \emph{Preprint}, arXiv:2307.13854.

\bibitem[{Zhu et~al.(2026)Zhu, Zhang, Ma, Xu, Zhang, Yang, Wang, Qiu, Wu, Dai,
  Ma, Liu, Yang, Luo, Yang, Li, Wang, Chen, Geng, and Guo}]{zhu2026retrac}
Jialiang Zhu, Gongrui Zhang, Xiaolong Ma, Lin Xu, Miaosen Zhang, Ruiqi Yang,
  Song Wang, Kai Qiu, Zhirong Wu, Qi~Dai, Ruichun Ma, Bei Liu, Yifan Yang,
  Chong Luo, Zhengyuan Yang, Linjie Li, Lijuan Wang, Weizhu Chen, Xin Geng, and
  Baining Guo. 2026.
\newblock \href {https://arxiv.org/abs/2602.02486} {Re-trac: Recursive
  trajectory compression for deep search agents}.
\newblock \emph{Preprint}, arXiv:2602.02486.

\bibitem[{Zhuge et~al.(2024)Zhuge, Wang, Kirsch, Faccio, Khizbullin, and
  Schmidhuber}]{zhuge2024gptswarm}
Mingchen Zhuge, Wenyi Wang, Louis Kirsch, Francesco Faccio, Dmitrii Khizbullin,
  and J{\"u}rgen Schmidhuber. 2024.
\newblock {GPTSwarm}: Language agents as optimizable graphs.
\newblock In \emph{Proceedings of the 41st International Conference on Machine
  Learning}.

\end{thebibliography}
\clearpage
\onecolumn
\renewcommand{\topfraction}{0.9}
\renewcommand{\bottomfraction}{0.7}
\renewcommand{\textfraction}{0.05}
\renewcommand{\floatpagefraction}{0.75}
\makeatletter
\setlength{\@fptop}{0pt}   %
\makeatother
\appendix
\section{Additional Construction Details}
\label{app:construction-details}

\subsection{Evidence Sources and Visit Attribution}

The construction pipeline uses two evidence channels. First, each think block is sentence-split into stable sentence identifiers so that query-side justifications can cite exact reasoning sentences. Second, earlier tool outputs are segmented into attributable provenance windows from search snippets and visited-page summaries. In the implementation, visited-page evidence is treated as stronger than snippet evidence, and snippet evidence is treated as stronger than think-only evidence.

Visited-page attribution is resolved by a deterministic priority order: exact URL match against same-turn search results, exact match in earlier turns, URL-topic fallback based on salient path segments, and finally $Prior\_knowledge$ fallback when no owner can be recovered. When ownership is recovered only through URL-topic fallback, the system also records an auxiliary $Prior\_knowledge$ source because the agent likely constructed the authoritative URL instead of directly retrieving it.

\subsection{Signal Statuses, Candidate Parents, and MPSC}

For each target query $q$, the implementation tokenizes the query into normalized signals and assigns each signal one of four statuses using only earlier turns as context: \qtag{Q0\_ONLY}, \qtag{Q0\_SEEN}, \qtag{SEEN}, and \qtag{NEW}. Signals marked \qtag{Q0\_ONLY} are treated as direct carry-over from the original question and do not require query-to-query parents. The remaining supported signals define the used-signal set $U(q)$.

Candidate parents may come only from prior queries that appear in provenance for signals in $U(q)$, from $Prior\_knowledge$ when an explicitly introduced signal lacks retrievable provenance, or from deterministic failure logic. Each candidate parent $p$ covers a subset of the target's used signals. Minimal parent-set cover (MPSC) retains a minimal covering parent set, implemented as a deterministic greedy set-cover approximation with recency-first tie-breaking and lowest priority for $Prior\_knowledge$ under equal gain. This keeps the query-side graph sparse and stable across runs. Recency-first tie-breaking reflects the serial-position behavior of LLMs in long contexts: when the same information appears at multiple positions, models rely most on its most recent occurrence~\citep{liu2024lost,guo2025serial}, so among earlier queries supplying the same clue, the latest copy is the one most plausibly conditioning the next query.

\subsection{Failure-Response Edges and Orphan Queries}

Failure transitions are encoded explicitly rather than being folded into topical similarity. Hard failures correspond to unusable retrieval such as zero-result pages, access denial, login walls, or HTTP failures. Soft failures correspond to explicit insufficiency statements in the current think block. The implementation always preserves deterministic soft anchor-to-anchor edges once a soft-failure pattern is detected, even if the next turn also receives evidence-use parents.

Two conservative exclusions prevent over-attribution. First, same-turn queries are never connected because they are parallel actions. Second, a query may remain orphaned when no admissible parent can be justified. The graph is therefore allowed to be disconnected at the query level when the trajectory itself does not expose a recoverable source.

\subsection{Answer-Support Frontier Details}
\label{app:answer-details}

The answer node preserves the original final answer text but is grounded through a smaller set of stable answer signals, including whole-answer spans, text-anchored answer units, and high-value names, numbers, years, and acronyms. A query receives a $q \rightarrow \mathrm{Answer}$ edge only when its attributable tool outputs support at least one answer signal. When no supporting query provenance exists for an answer
signal, a $Prior\_knowledge \rightarrow \mathrm{Answer}$ edge is added instead, indicating that the answer is unsupported by retrieval and treated as drawn from the model's prior knowledge.

Answer-side parent selection follows a frontier policy instead of strict minimality. The system retains strong support candidates first, then adds extra parents only when necessary to cover still-unsupported answer signals. It prunes only locally dominated ancestors from the same provenance cluster and performs light deduplication afterward. This yields a sparse frontier while preserving multiple supporting branches when they genuinely contribute distinct answer evidence.

\FloatBarrier

\FloatBarrier

\subsection{Representative DAG Overview}
\label{app:representative-dag}

The BrowseComp portion of the 100-DAG canonical validation package contains 50 DAGs: ten tasks (BC7, BC9, BC10, BC11, BC12, BC15, BC22, BC33, BC40, BC44) crossed with five systems. This subset provides the detailed worked examples used for illustration, while Appendix~\ref{app:DAG_parsing_validation} reports the full validation package across BrowseComp, WebWalker-Hard, and DeepSearchQA.

\begin{figure}[htbp]
  \centering
  \includegraphics[width=0.98\textwidth]{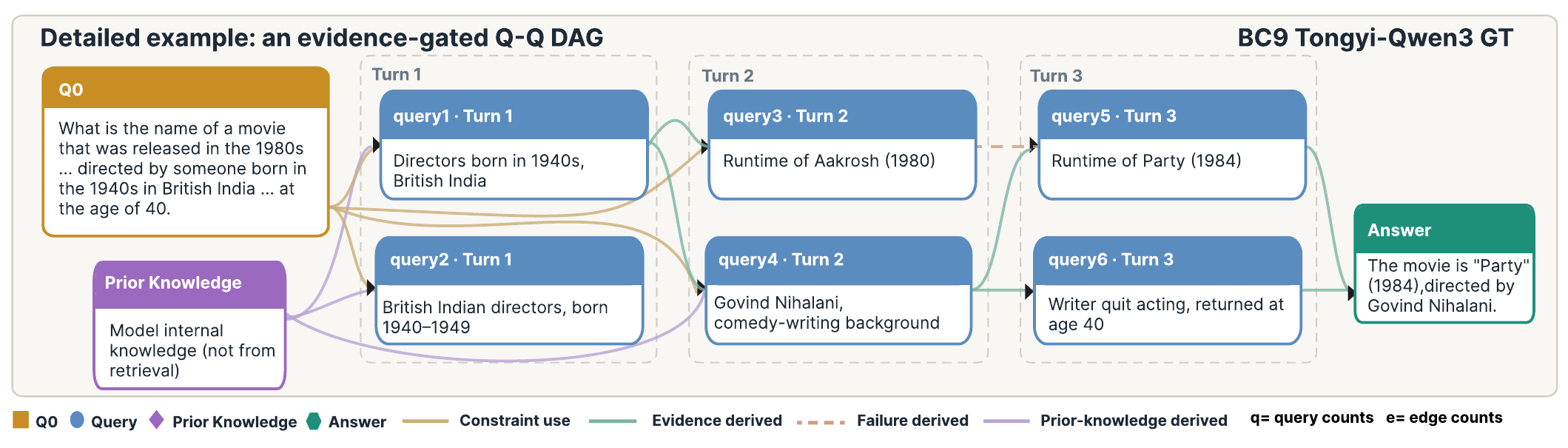}
  \caption{Worked example of a \textsc{SearchAtlas} DAG for one trajectory. The figure shows how query nodes, evidence-use query-to-query edges, prior-knowledge attribution, and answer-support edges are represented in a concrete case.}
  \label{fig:detailed-example}
\end{figure}

\FloatBarrier

\subsection{DAG Parsing Validation}
\label{app:DAG_parsing_validation}

Human-labeled DAGs serve as canonical references for evaluating the generated DAG pipeline. Two annotators independently labeled each trajectory using the same raw trajectory evidence and the guidelines summarized below. Edges on which the annotators agreed formed an initial consensus graph; remaining ambiguous or disputed edges were reviewed against the raw trajectory and adjudicated through a five-author audit to produce the final canonical DAG. The annotation target is the model's actual trajectory and final answer, not the dataset's gold answer.

\begin{figure}[htbp]
  \centering
  \IfFileExists{figures/human_anno.png}{%
    \includegraphics[width=\linewidth]{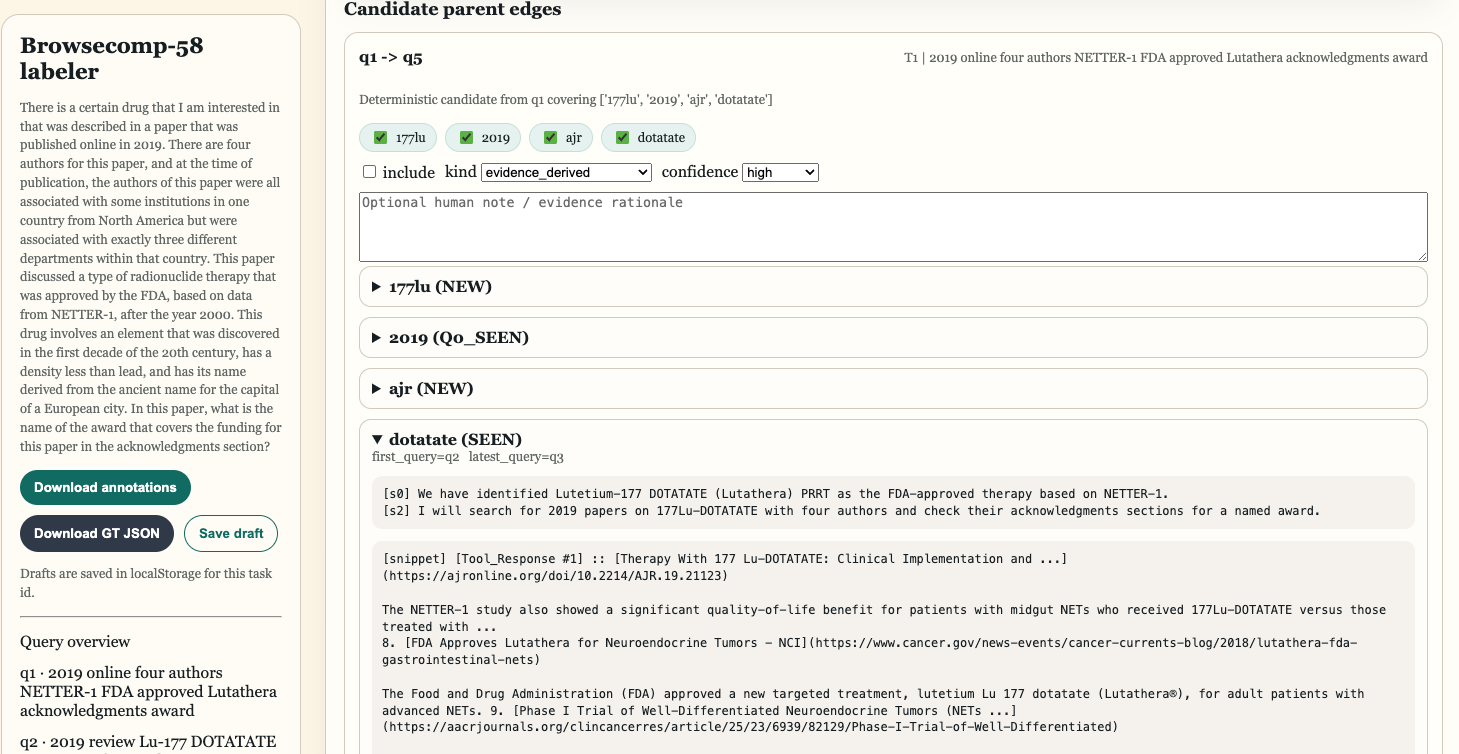}
  }{%
    \fbox{\parbox{0.92\linewidth}{\centering Placeholder for the human DAG annotation interface screenshot.}}
  }
  \caption{Human DAG annotation interface used to construct canonical DAGs. The interface lets annotators review each candidate parent query and inspect the evidence provenance of individual query tokens.}
  \label{fig:human_annotation}
\end{figure}

\paragraph{Annotation inputs and workflow.}
Annotators used an interactive HTML labeling tool organized query by query (Figure~\ref{fig:human_annotation}). The interface presented the original question, the model's final answer, automatically generated deterministic \qzero edges, and the sequence of extracted search queries. For each query, annotators first reviewed the query content, the reasoning block that issued it, the retrieved tool results, visited webpages, and a word-status table labeling each query word as copied from the original question, previously seen, introduced from earlier retrieved evidence, or unsupported; for any word whose provenance was not established by exact carryover, they used this evidence to classify it as a \qzero paraphrase, an unsupported prior-knowledge contribution, or unresolved.

Second, for each query, the interface displayed candidate parent queries, edge type, confidence, notes, and the words of that query covered by that candidate parent. For each covered signal, the tool exposed supporting evidence such as first and latest query occurrence, matching reasoning sentences, and provenance windows from source queries. Finally, annotators reviewed answer-support candidates from query nodes to the final answer and exported the resulting graph JSON.

Table~\ref{tab:human-annotation-rules} summarizes the edge-specific annotation rules. Annotators add an edge only when they can identify visible evidence, a visible failure, an original-question constraint, or an unsupported model assumption that explains the target query or answer. 
\begin{table}[htbp]
\centering
\footnotesize
\setlength{\tabcolsep}{4pt}
\begin{tabularx}{\linewidth}{@{}p{0.22\linewidth}YY@{}}
\toprule
Edge type & Use when & Conservative exclusion \\
\midrule
\qzero$\rightarrow q$ & The query directly uses a requirement from the original question. & Do not add a query--query edge for words already explained by \qzero. \\
$q\rightarrow q$ & Retrieved snippets or visited pages supply a concrete clue used by a later query. & Do not connect broad topical overlap or chronology alone. \\
$q\rightarrow A$ & Retrieved evidence supports a factual unit in the answer. & Do not force unsupported answer parts onto a query. \\
PK$\rightarrow q/A$ & The model introduces a concrete unsupported assumption, guess, calculation, or constructed URL. & Do not use PK when retrieved evidence supports the same signal. \\
failure-response edge & A later query is explicitly linked to a failed or insufficient earlier search. & Do not add failure-response edges merely because the final answer is wrong. \\
\bottomrule
\end{tabularx}
\caption{Condensed human annotation rules for canonical DAG construction. The full internal guideline uses the same principle throughout: visible contribution, not possible relevance.}
\label{tab:human-annotation-rules}
\end{table}

\paragraph{Human disagreement and adjudication.}
Table~\ref{tab:annotation-disagreements} summarizes the recurring disagreement categories, and Table~\ref{tab:annotation-audit-examples} provides representative adjudication decisions from 39 trajectories. These tables document the annotation protocol and are not treated as additional independent labels.
\begin{table}[htbp]
\centering
\footnotesize
\setlength{\tabcolsep}{4pt}
\begin{tabularx}{\linewidth}{@{}p{0.25\linewidth}YY@{}}
\toprule
Disagreement type & Typical source & Adjudication rule \\
\midrule
Query parent attribution & Multiple earlier queries mention a related entity or source. & Keep the parent that supplies the concrete item reused by the target query. \\
PK attribution & A bridge term, year, entity, or URL may be inferred rather than retrieved. & Mark PK only when the signal is not traceable to \qzero or retrieved evidence. \\
Failure-response edge & A retry may follow a weak result without an explicit failure statement. & Add a failure-response edge only when failure or insufficiency is visible in the trajectory. \\
Visit provenance owner & A visited page may match several prior queries. & Assign the visit to the query whose result or topic most directly led to the page. \\
Answer grounding & Different queries support different answer spans. & Add $q\rightarrow A$ only for factual units in answer supported by retrieved evidence. \\
Parent-set redundancy & Several candidate parents explain the same target content. & Retain the minimal parent set that explains the grounded content. \\
\bottomrule
\end{tabularx}
\caption{Disagreement taxonomy used during adjudication of the canonical DAG annotations.}
\label{tab:annotation-disagreements}
\end{table}

\begin{table}[htbp]
\centering
\footnotesize
\setlength{\tabcolsep}{5pt}
\begin{tabularx}{\textwidth}{@{}p{0.16\textwidth}p{0.29\textwidth}p{0.24\textwidth}Y@{}}
\toprule
Audit issue & Representative case & Adjudicated decision & Rule illustrated \\
\midrule
Unsupported inferred signal & BC7 / TYDP: a later query introduced a computed year such as ``2012'' from reasoning rather than retrieval. & Replace weak query-parent attribution with $Prior\_knowledge$$\rightarrow q$. & PK attribution requires traceable retrieval support; unsupported arithmetic or guesses are PK. \\
Hard-failure retry & BC7 / TYDP: a zero-result query was immediately reformulated. & Add a failure-response edge from the failed query to the retry. & Failure-response edges require visible search failure or explicit insufficiency. \\
Soft-failure overuse & BC7 / TYDP: later turns continued after progress rather than after a stated insufficiency. & Remove soft failure-response edges. & Progression to the next clue is not itself a failure. \\
Visit provenance owner & BC15 / WebSailor: a visited MDPI page appeared in the same-turn search results for q19. & Assign the visit evidence to q19 rather than creating a visit node. & Visits are provenance owned by the query that led to the page. \\
Answer grounding & BC11 / WebSailor: the final answer ``Nick'' was supported by a visited page owned by q5. & Add q5$\rightarrow$Answer and remove orphan PK answer support. & The Answer node tracks the model's produced answer and should be grounded to retrieved support when available. \\
Generic-token parent noise & BC44 / TYDP-GPT5: edges were created from generic tokens such as \texttt{site}, \texttt{ac}, \texttt{uk}, or \texttt{profile}. & Delete generic-token edges and keep entity-specific parents. & Query-parent edges require concrete information gain, not search-operator overlap. \\
\bottomrule
\end{tabularx}
\caption{Examples of adjudication decisions drawn from the audited annotation reports. These examples instantiate the disagreement taxonomy in \autoref{tab:annotation-disagreements}.}
\label{tab:annotation-audit-examples}
\end{table}

\paragraph{Canonical annotation package scale.}
The 100 canonical DAGs are compact in number but not in annotation volume. As summarized in Table~\ref{tab:canonical-package-scale}, they cover three benchmarks and all five systems, requiring thousands of node and edge decisions.

\begin{table}[htbp]
\centering
\footnotesize
\setlength{\tabcolsep}{5pt}
\begin{tabularx}{\textwidth}{@{}p{0.25\textwidth}p{0.30\textwidth}Y@{}}
\toprule
Quantity & Value & Why it matters \\
\midrule
Canonical DAGs & 100 & BrowseComp, WebWalker-Hard, and DeepSearchQA examples crossed with five systems. \\
Annotated query nodes & 4,109 total & The validation set covers both compact and long search trajectories. \\
Temporal parent-pair search space & 281,460 possible \qzero/previous-query parent pairs before provenance filtering & Edge labels are evaluated over a much larger temporal search space than the number of final DAGs suggests. \\
Unique gold edge pairs & 7,432 total; mean 74.3 & Provides the edge-level support behind the reported graph $F_1$, which uses exact $(\mathrm{source},\mathrm{target})$ matches. \\
Unique gold edge pairs by type & evidence-use 5,310 (71.4\%); constraint-use 823 (11.1\%); prior-knowledge 656 (8.8\%); failure-response 643 (8.7\%) & Shows that validation covers all edge classes used by the pipeline. \\
\bottomrule
\end{tabularx}

\vspace{0.5em}
\begin{tabular}{@{}lrrr@{}}
\toprule
Model & DAGs & Query nodes mean / median & Gold edge-pair mean \\
\midrule
TYDP & 20 & 42.8 / 33.0 & 80.4 \\
MiroThinker & 20 & 122.4 / 110.5 & 214.1 \\
TYDP-GPT5 & 20 & 15.7 / 13.5 & 31.0 \\
TYDP-Qwen3 & 20 & 8.6 / 3.0 & 14.9 \\
WebSailor & 20 & 16.1 / 19.0 & 31.3 \\
\bottomrule
\end{tabular}

\caption{Scale of the human canonical DAG validation package. The parent-pair count is the full temporal search space implied by query order. For each query, \qzero and all previous queries are possible parents before provenance filtering. It is not the number of manually clicked include/exclude decisions.}
\label{tab:canonical-package-scale}
\end{table}

\FloatBarrier

These tables are the appendix counterpart of the reliability claim in Section~3. They summarize the stability and ground-truth-aligned (GT-aligned) graph-accuracy results for the 100 canonical DAG package: ten BrowseComp tasks and five tasks each from WebWalker-Hard and DeepSearchQA, all crossed with five systems. Graph precision, recall, and $F_1$ compare each majority-vote graph against its human-adjudicated canonical DAG using exact $(\mathrm{source}, \mathrm{target})$ edge matches. Each metric is computed independently per DAG and then macro-averaged separately; consequently, the reported mean $F_1$ is not obtained by taking the harmonic mean of the reported mean precision and recall. Stability is the mean pairwise Jaccard over repeated generated DAGs. Seventy-one of the 100 canonical DAGs contain at least one failure-response relation, illustrating that recovery and failed branches are common in the validation set.

\begin{table}[htbp]
\centering
\small
\setlength{\tabcolsep}{4pt}
\begin{tabular}{llrrrrr}
\toprule
Dataset & Model & DAGs & Precision & Recall & Graph $F_1$ & Stability \\
\midrule
BrowseComp & MiroThinker & 10 & 0.897 & 0.945 & 0.911 & 0.825 \\
BrowseComp & TYDP & 10 & 0.834 & 0.887 & 0.858 & 0.820 \\
BrowseComp & TYDP-GPT5 & 10 & 0.809 & 0.900 & 0.846 & 0.855 \\
BrowseComp & TYDP-Qwen3 & 10 & 0.890 & 0.840 & 0.861 & 0.954 \\
BrowseComp & WebSailor & 10 & 0.868 & 0.893 & 0.878 & 0.796 \\
\midrule
WebWalker-Hard & MiroThinker & 5 & 1.000 & 0.664 & 0.786 & 0.891 \\
WebWalker-Hard & TYDP & 5 & 0.972 & 0.764 & 0.845 & 0.879 \\
WebWalker-Hard & TYDP-GPT5 & 5 & 0.975 & 0.871 & 0.904 & 0.962 \\
WebWalker-Hard & TYDP-Qwen3 & 5 & 1.000 & 0.878 & 0.934 & 1.000 \\
WebWalker-Hard & WebSailor & 5 & 1.000 & 0.835 & 0.891 & 0.845 \\
\midrule
DeepSearchQA & MiroThinker & 5 & 0.854 & 0.775 & 0.795 & 0.803 \\
DeepSearchQA & TYDP & 5 & 0.817 & 0.902 & 0.855 & 0.795 \\
DeepSearchQA & TYDP-GPT5 & 5 & 0.900 & 0.821 & 0.844 & 0.937 \\
DeepSearchQA & TYDP-Qwen3 & 5 & 0.946 & 0.742 & 0.814 & 0.849 \\
DeepSearchQA & WebSailor & 5 & 0.833 & 0.862 & 0.835 & 0.937 \\
\bottomrule
\end{tabular}
\caption{GT-aligned graph reconstruction and run stability for the 100 human-adjudicated canonical DAGs. Precision, recall, and Graph $F_1$ are computed per DAG and macro-averaged separately within each dataset--model group.}
\label{tab:stability-accuracy-100dags}
\end{table}

\begin{table}[htbp]
\centering
\small
\begin{tabular}{lrrrrr}
\toprule
Dataset & DAGs & Precision & Recall & Graph $F_1$ & Stability \\
\midrule
BrowseComp & 50 & 0.859 & 0.893 & 0.871 & 0.850 \\
WebWalker-Hard & 25 & 0.989 & 0.802 & 0.872 & 0.915 \\
DeepSearchQA & 25 & 0.870 & 0.821 & 0.828 & 0.864 \\
\midrule
All & 100 & 0.895 & 0.852 & 0.860 & 0.870 \\
\bottomrule
\end{tabular}
\caption{Dataset-level reliability summary for the human-adjudicated canonical package. Precision, recall, and Graph $F_1$ are computed per DAG and macro-averaged separately.}
\label{tab:stability-accuracy-100dags-dataset}
\end{table}

\begin{table}[htbp]
\centering
\small
\begin{tabular}{lrrrr}
\toprule
Edge type & Precision & Recall & $F_1$ & Stability \\
\midrule
Constraint-use & 0.958 & 0.979 & 0.957 & 1.000 \\
Evidence-use & 0.797 & 0.791 & 0.769 & 0.795 \\
Prior-knowledge & 0.813 & 0.784 & 0.758 & 0.963 \\
Failure-response & 0.871 & 0.715 & 0.759 & 0.982 \\
\bottomrule
\end{tabular}
\caption{Edge-type-level GT alignment and repeat stability for the 100 canonical DAG package. For each type, precision, recall, and $F_1$ use exact source--target matches within each DAG and are macro-averaged separately; a type absent from both graphs is treated as an exact match. Stability is pairwise Jaccard computed within each edge type before averaging across DAGs.}
\label{tab:edge-type-reliability-100dags}
\end{table}

\begin{table}[htbp]
\centering
\small
\begin{tabular}{lrrrrr}
\toprule
Dataset & DAGs & Nodes & Query nodes & Unique gold edge pairs & Candidate parent pairs \\
\midrule
BrowseComp & 50 & 2919 & 2769 & 4950 & 237908 \\
WebWalker-Hard & 25 & 360 & 285 & 547 & 2467 \\
DeepSearchQA & 25 & 1130 & 1055 & 1935 & 41085 \\
\midrule
All & 100 & 4409 & 4109 & 7432 & 281460 \\
\bottomrule
\end{tabular}
\caption{Scale of the human-adjudicated canonical DAG package. Gold edges are unique $(\mathrm{source},\mathrm{target})$ pairs, matching the evaluation unit. Candidate parent pairs count possible earlier-query parent choices for each query node.}
\label{tab:canonical-gt-scale-100dags}
\end{table}

\FloatBarrier

We test whether reconstruction and downstream conclusions depend on the LLM used for local attribution. We hold the trajectory inputs, graph ontology, deterministic preprocessing, and evaluation protocol fixed, and replace only the attribution model. Table~\ref{tab:constructor-canonical-sensitivity} evaluates four constructors against the same 100 human-adjudicated DAGs. Precision, recall, and Graph $F_1$ are computed per DAG and macro-averaged separately. All four constructors retain substantial agreement with human annotations, with Graph $F_1$ between $0.814$ and $0.860$.

\begin{table}[htbp]
\centering
\small
\setlength{\tabcolsep}{10pt}
\begin{tabular}{lrrrr}
\toprule
DAG constructor & $N$ & Precision & Recall & Graph $F_1$ \\
\midrule
GPT-5.2 & 100 & 0.895 & 0.852 & 0.860 \\
GLM-5.2 & 100 & 0.810 & 0.894 & 0.846 \\
Gemini-3.5-Flash & 100 & 0.784 & 0.855 & 0.814 \\
Claude-4.5-Haiku & 100 & 0.808 & 0.887 & 0.842 \\
\bottomrule
\end{tabular}
\caption{Sensitivity of canonical-DAG reconstruction to the local attribution model. All non-constructor components are held fixed.}
\label{tab:constructor-canonical-sensitivity}
\end{table}

At full-set scale, we reconstruct the 1,350-trajectory evaluation with GLM-5.2 and Gemini-3.5-Flash. Relative to the GPT-5.2 graphs, their per-trajectory macro edge $F_1$ scores are $0.866$ and $0.870$, respectively. We then recompute the complete question-held-out diagnostic evaluation. Table~\ref{tab:constructor-downstream-sensitivity} shows that all constructor--regime comparisons preserve the same stage-wise result: constraint grounding improves on topology alone, and PK reliance provides a further improvement. Final AUCs remain between $0.760$ and $0.861$.

\begin{table}[htbp]
\centering
\small
\setlength{\tabcolsep}{12pt}
\begin{tabular}{lccc}
\toprule
Dataset / regime & GPT-5.2 & GLM-5.2 & Gemini-3.5-Flash \\
\midrule
BrowseComp / sequential & 0.714 / 0.816 / 0.855 & 0.699 / 0.791 / 0.831 & 0.699 / 0.786 / 0.823 \\
WebWalker-Hard / parallel & 0.746 / 0.779 / 0.840 & 0.724 / 0.745 / 0.785 & 0.735 / 0.760 / 0.819 \\
DeepSearchQA / sequential & 0.734 / 0.840 / 0.844 & 0.729 / 0.793 / 0.827 & 0.671 / 0.752 / 0.760 \\
DeepSearchQA / parallel & 0.771 / 0.830 / 0.856 & 0.782 / 0.787 / 0.861 & 0.753 / 0.777 / 0.803 \\
\bottomrule
\end{tabular}
\caption{Stage-wise held-out AUC under different DAG constructors. Each entry reports Topology / + Grounding / + PK-risk penalty.}
\label{tab:constructor-downstream-sensitivity}
\end{table}

We also report the inference cost of graph construction. 

\begin{table}[htbp]
\centering
\small
\setlength{\tabcolsep}{7pt}
\begin{tabular}{lrrrl}
\toprule
DAG constructor & Input / output per 1M tokens & 1,350 DAGs & Per DAG \\
\midrule
Gemini-3.5-Flash & \$1.50 / \$9.00 & $\sim$\$2,000 & $\sim$\$1.48 \\
GPT-5.2 & \$1.75 / \$14.00 & $\sim$\$2,540 & $\sim$\$1.88 \\
Claude-4.5-Haiku & \$1.00 / \$5.00 & $\sim$\$1,270 & $\sim$\$0.94 \\
\bottomrule
\end{tabular}
\caption{Graph-construction cost for the 1,350-trajectory evaluation.}
\label{tab:constructor-cost}
\end{table}

\FloatBarrier

The contrast with direct one-shot construction, which reaches only $0.072$ edge $F_1$ and $0.455$ repeat Jaccard, indicates that the structured harness---deterministic preprocessing, local evidence attribution, and parent-set pruning---is central to this robustness.

We also test whether the graph can be recovered by prompting GPT-5.2 to directly produce a DAG from the trajectory, without the structured construction pipeline. The pilot covers the 50-DAG BrowseComp subset of the canonical package: ten BrowseComp questions, five agents, and three repeated direct-generation runs per case. We compare each generated graph against the human-adjudicated canonical graph using exact source--target edge-pair precision, recall, and $F_1$; edge kinds are ignored in this comparison. Repeat stability is measured by repeat-to-repeat edge Jaccard.

\begin{promptbox}{Direct GPT-5.2 DAG Baseline Prompt}
\textbf{Role.} Convert a raw web-search / deep-research trajectory directly into a DAG using the same ontology as the staged Query-DAG pipeline.

\textbf{Input.} The model receives raw trajectory text, including the original question, issued queries/actions, tool results when present, reasoning text, and final answer.

\textbf{Nodes.} Include exactly one Q0 node, one chronological query node for each distinct search/query/action request, exactly one Answer node, and a Prior\_knowledge node only if an edge uses it.

\textbf{Answer units.} Decompose the provided final answer into short answer units. Each answer unit must be a literal or normalized span from the final answer.

\textbf{Edge kinds.} Use Q0$\rightarrow$q constraint-use edges when a query deploys a question constraint; q$\rightarrow$q or q$\rightarrow$Answer evidence-use edges only when the trajectory shows that retrieved evidence was materially used; Prior\_knowledge$\rightarrow$q/Answer prior-knowledge edges only when a necessary signal is introduced without trajectory evidence; and failure-response edges only when the trajectory explicitly states a failed or insufficient earlier query and the later query repairs that gap.

\textbf{Hard rules.} Do not use outside knowledge, correct the agent, infer hidden tool results, hallucinate evidence, create Answer support from topical overlap, or add redundant edges. If support is uncertain, omit the edge or mark confidence as uncertain.

\textbf{Output.} Return JSON only with \texttt{q0\_units}, \texttt{answer\_units}, graph nodes, graph edges, \texttt{no\_source\_found}, and brief notes.
\end{promptbox}

\begin{table}[htbp]
\centering
\small
\setlength{\tabcolsep}{4pt}
\begin{tabular}{rrrrrr}
\toprule
Runs & Cases & Precision & Recall & $F_1$ & Repeat Jaccard \\
\midrule
150 & 50 & 0.497 & 0.039 & 0.072 & 0.455 \\
\bottomrule
\end{tabular}
\caption{Overall direct GPT-5.2 DAG-construction baseline. Direct generation obtains moderate precision but extremely low recall and $F_1$, indicating that it misses most GT edges.}
\label{tab:direct-gpt-overall}
\end{table}

\begin{table}[htbp]
\centering
\small
\setlength{\tabcolsep}{3pt}
\begin{minipage}[t]{0.48\linewidth}
\centering
\textbf{By agent}\\[2pt]
\begin{tabular}{lrrrrr}
\toprule
Agent & Runs & P & R & $F_1$ & Jac. \\
\midrule
TYDP           & 30 & 0.424 & 0.051 & 0.092 & 0.288 \\
MiroThinker    & 30 & 0.344 & 0.006 & 0.013 & 0.290 \\
TYDP-GPT5      & 30 & 0.634 & 0.130 & 0.215 & 0.528 \\
TYDP-Qwen3     & 30 & 0.526 & 0.326 & 0.403 & 0.520 \\
WebSailor      & 30 & 0.544 & 0.123 & 0.200 & 0.650 \\
\bottomrule
\end{tabular}
\end{minipage}\hfill
\begin{minipage}[t]{0.48\linewidth}
\centering
\textbf{By question}\\[2pt]
\begin{tabular}{lrrrrr}
\toprule
Question & Runs & P & R & $F_1$ & Jac. \\
\midrule
BC7  & 15 & 0.533 & 0.024 & 0.046 & 0.289 \\
BC9  & 15 & 0.381 & 0.032 & 0.058 & 0.362 \\
BC10 & 15 & 0.402 & 0.033 & 0.061 & 0.395 \\
BC11 & 15 & 0.524 & 0.020 & 0.038 & 0.660 \\
BC12 & 15 & 0.691 & 0.035 & 0.067 & 0.461 \\
BC15 & 15 & 0.551 & 0.187 & 0.280 & 0.514 \\
BC22 & 15 & 0.606 & 0.083 & 0.147 & 0.526 \\
BC33 & 15 & 0.590 & 0.029 & 0.055 & 0.383 \\
BC40 & 15 & 0.484 & 0.067 & 0.118 & 0.502 \\
BC44 & 15 & 0.403 & 0.062 & 0.107 & 0.457 \\
\bottomrule
\end{tabular}
\end{minipage}
\caption{Direct GPT-5.2 DAG baseline by agent and by question. P, R, $F_1$, and Jac. denote micro precision, micro recall, micro $F_1$, and repeat-to-repeat edge Jaccard. Recall remains low across both agents and validation questions, showing that direct prompting is not a substitute for the structured attribution pipeline.}
\label{tab:direct-gpt-breakdown}
\end{table}

The direct baseline usually emits sparse graphs that contain a small number of plausible edges, which explains the moderate precision, but it fails to recover the dense attribution structure needed for query-to-query analysis. The result motivates the design choice in Section~3: the LLM is used for narrow attribution decisions inside a constrained pipeline, rather than being asked to synthesize the entire DAG in one step.

\FloatBarrier
\subsection{Human Audit of Question Type and Constraint Decomposition}
\label{app:annotation-audit-results}

We separately audit the two annotations used by the downstream diagnostics. Constraint units are proposed by an LLM, checked by a deterministic parser, and independently reviewed by two annotators; question types are reviewed under the operational dependency definition in Appendix~\ref{app:question-type-prompt}. For each annotation target, agreed labels form the initial consensus and disagreements are resolved through five-annotator adjudication. Table~\ref{tab:question-constraint-audit} reports exact agreement before adjudication.

\begin{table}[htbp]
\centering
\small
\setlength{\tabcolsep}{8pt}
\begin{tabular}{lrrr}
\toprule
Annotation target & Total & Disagreements & Exact agreement \\
\midrule
Question type & 270 & 5 & 98.15\% \\
Question-level decomposition & 270 & 6 & 97.78\% \\
Aligned constraint units & 1,685 & 39 & 97.69\% \\
\bottomrule
\end{tabular}
\caption{Independent human audit of question-type and constraint annotations before adjudication.}
\label{tab:question-constraint-audit}
\end{table}
\FloatBarrier
\subsection{Question-Type Classification Prompt}
\label{app:question-type-prompt}

The following prompt follows the definitions in Section~\ref{sec:process-diagnostics}. It treats dependency-carrying search, ranked or survivor-set maintenance, and derived-owner mappings as sequential constraints, and direct-field evaluation over a shared target or bounded candidate space as parallel constraints.

\begin{promptbox}{Question-Type Classification Prompt}
Classify each question as exactly one of:

\textbf{SEQUENTIAL\_CONSTRAINT:}
The question requires dependency-carrying search. The solver must first construct or bind an intermediate candidate set, entity, source, owner, ranking, or eligibility state, and later constraints depend on that intermediate state before the final answer can be selected.

\textbf{PARALLEL\_CONSTRAINT:}
The question requires direct candidate evaluation. The constraints are direct predicates on the same target variable, such as attributes, dates, counts, rankings, memberships, thresholds, comparisons, semantic matches, or simple aggregations over a shared candidate space.

\textbf{Core distinction.}
Do not classify by the number of lookups, sources, or constraints. If all conditions can be checked as direct predicates on the same candidate $x$, choose \textbf{PARALLEL\_CONSTRAINT}. Choose \textbf{SEQUENTIAL\_CONSTRAINT} when earlier retrieval changes what later constraints are evaluated against, or when a constructed candidate state must be preserved through downstream checks.

\textbf{Decision rules.}
\begin{enumerate}
\item \textbf{Direct-field rule:} Use \textbf{PARALLEL\_CONSTRAINT} when the question asks for an entity or direct attribute and all constraints describe direct fields of that same target.
\item \textbf{Compact-anchor rule:} Use \textbf{PARALLEL\_CONSTRAINT} when the question is anchored to a named table, report, website, organization, event corpus, legal context, roster, or bounded candidate set, and the remaining work only reads, filters, compares, intersects, or aggregates direct attributes of those candidates.
\item \textbf{Dependency-chain rule:} Use \textbf{SEQUENTIAL\_CONSTRAINT} when the solver must first resolve an intermediate entity, value, source, owner, or candidate set before later constraints become meaningful.
\item \textbf{Ranked/survivor-state rule:} Use \textbf{SEQUENTIAL\_CONSTRAINT} when a top-$k$, bottom-$k$, highest/lowest group, shortlist, or survivor set is constructed and then carried through multiple downstream checks before the final answer is selected.
\item \textbf{Derived-owner rule:} Use \textbf{SEQUENTIAL\_CONSTRAINT} when a candidate must be mapped to another evidence target before the decisive check, such as entity $\rightarrow$ source, university $\rightarrow$ city, event $\rightarrow$ date/month, person $\rightarrow$ work, item $\rightarrow$ linked document, or legal/textual item $\rightarrow$ cited item.
\item \textbf{Broad-survivor rule:} Use \textbf{SEQUENTIAL\_CONSTRAINT} for broad research-style questions where the answer is an obscure survivor of many heterogeneous public-data, institutional, technical, or source-specific constraints.
\end{enumerate}

\textbf{Conflict resolution.}
Prefer \textbf{PARALLEL\_CONSTRAINT} for direct target-field evaluation over a shared candidate space. Prefer \textbf{SEQUENTIAL\_CONSTRAINT} when the question requires maintaining an intermediate state, transferring a ranked or survivor set, mapping to a derived owner/source, or resolving a dependency path before the final answer can be selected.

\textbf{Few-shot boundaries.}

\textit{Example 1.}
Question pattern: A top-five candidate set is first obtained from one source, then carried into later checks or mapped to another owner/source before choosing the final answer.
Label: \textbf{SEQUENTIAL\_CONSTRAINT}.
Reason: the top-five set is an intermediate search state that must be preserved through downstream evidence.

\textit{Example 2.}
Question pattern: A named source, table, report, or compact candidate set is given, and each candidate is checked against direct fields; or two small lists are intersected and one direct field is read.
Label: \textbf{PARALLEL\_CONSTRAINT}.
Reason: each condition is a direct predicate on the same candidate variable.

\textbf{Return format.}
Return exactly one JSON object with two fields: \texttt{"category"}, whose value is either \texttt{"SEQUENTIAL\_CONSTRAINT"} or \texttt{"PARALLEL\_CONSTRAINT"}, and \texttt{"reason"}, a 1--3 sentence explanation.
\end{promptbox}

\subsection{Benchmark Subset Question IDs}
\label{app:question-id-list}

We report the exact question subsets used in the main experiments in \autoref{tab:benchmark-question-ids}. The subsets are selected after applying the question-type classifier in Appendix~\ref{app:question-type-prompt}: BrowseComp is used as the sequential source because almost all questions in the full pool are classified as sequential, while WebWalker-Hard-English is used as the parallel source because almost all questions in the hard-English split are classified as parallel. For BrowseComp, we remove geography-only direct-field items and take the first 150 remaining questions by official ID. For DeepSearchQA, which contains both regimes, we sample 25 classified sequential questions and 25 classified parallel questions.

\begin{table}[htbp]
\centering
\scriptsize
\setlength{\tabcolsep}{4pt}
\begin{tabularx}{\textwidth}{@{}p{0.20\textwidth}p{0.13\textwidth}Y@{}}
\toprule
Benchmark subset & Regime & Question IDs \\
\midrule
BrowseComp & Sequential &
BrowseComp-1, BrowseComp-3, BrowseComp-6, BrowseComp-7, BrowseComp-8, BrowseComp-9, BrowseComp-10, BrowseComp-11, BrowseComp-12, BrowseComp-13, BrowseComp-15, BrowseComp-17, BrowseComp-18, BrowseComp-19, BrowseComp-20\newline
BrowseComp-21, BrowseComp-22, BrowseComp-23, BrowseComp-24, BrowseComp-25, BrowseComp-26, BrowseComp-27, BrowseComp-28, BrowseComp-29, BrowseComp-30, BrowseComp-31, BrowseComp-32, BrowseComp-33, BrowseComp-36, BrowseComp-37\newline
BrowseComp-38, BrowseComp-39, BrowseComp-40, BrowseComp-41, BrowseComp-42, BrowseComp-43, BrowseComp-44, BrowseComp-46, BrowseComp-47, BrowseComp-48, BrowseComp-49, BrowseComp-50, BrowseComp-51, BrowseComp-52, BrowseComp-54\newline
BrowseComp-55, BrowseComp-56, BrowseComp-57, BrowseComp-58, BrowseComp-59, BrowseComp-60, BrowseComp-61, BrowseComp-62, BrowseComp-63, BrowseComp-64, BrowseComp-65, BrowseComp-66, BrowseComp-67, BrowseComp-68, BrowseComp-69\newline
BrowseComp-70, BrowseComp-73, BrowseComp-75, BrowseComp-76, BrowseComp-77, BrowseComp-78, BrowseComp-79, BrowseComp-80, BrowseComp-81, BrowseComp-83, BrowseComp-85, BrowseComp-86, BrowseComp-89, BrowseComp-90, BrowseComp-91\newline
BrowseComp-92, BrowseComp-93, BrowseComp-94, BrowseComp-95, BrowseComp-97, BrowseComp-99, BrowseComp-100, BrowseComp-101, BrowseComp-102, BrowseComp-103, BrowseComp-104, BrowseComp-105, BrowseComp-106, BrowseComp-107, BrowseComp-108\newline
BrowseComp-109, BrowseComp-110, BrowseComp-111, BrowseComp-112, BrowseComp-113, BrowseComp-114, BrowseComp-115, BrowseComp-116, BrowseComp-117, BrowseComp-118, BrowseComp-119, BrowseComp-120, BrowseComp-121, BrowseComp-123, BrowseComp-124\newline
BrowseComp-125, BrowseComp-126, BrowseComp-127, BrowseComp-128, BrowseComp-129, BrowseComp-130, BrowseComp-131, BrowseComp-132, BrowseComp-134, BrowseComp-136, BrowseComp-139, BrowseComp-141, BrowseComp-142, BrowseComp-143, BrowseComp-145\newline
BrowseComp-146, BrowseComp-147, BrowseComp-149, BrowseComp-150, BrowseComp-151, BrowseComp-152, BrowseComp-153, BrowseComp-154, BrowseComp-155, BrowseComp-156, BrowseComp-157, BrowseComp-158, BrowseComp-159, BrowseComp-160, BrowseComp-161\newline
BrowseComp-162, BrowseComp-164, BrowseComp-165, BrowseComp-166, BrowseComp-167, BrowseComp-168, BrowseComp-169, BrowseComp-170, BrowseComp-171, BrowseComp-173, BrowseComp-174, BrowseComp-175, BrowseComp-176, BrowseComp-178, BrowseComp-180 \\
\midrule
WebWalker-Hard-English & Parallel &
WebWalkerQA-Hard-002, WebWalkerQA-Hard-003, WebWalkerQA-Hard-004, WebWalkerQA-Hard-005, WebWalkerQA-Hard-007, WebWalkerQA-Hard-009, WebWalkerQA-Hard-010, WebWalkerQA-Hard-011, WebWalkerQA-Hard-012, WebWalkerQA-Hard-014, WebWalkerQA-Hard-015, WebWalkerQA-Hard-016, WebWalkerQA-Hard-017, WebWalkerQA-Hard-018, WebWalkerQA-Hard-019\newline
WebWalkerQA-Hard-021, WebWalkerQA-Hard-022, WebWalkerQA-Hard-023, WebWalkerQA-Hard-024, WebWalkerQA-Hard-025, WebWalkerQA-Hard-026, WebWalkerQA-Hard-027, WebWalkerQA-Hard-028, WebWalkerQA-Hard-029, WebWalkerQA-Hard-030, WebWalkerQA-Hard-031, WebWalkerQA-Hard-032, WebWalkerQA-Hard-033, WebWalkerQA-Hard-034, WebWalkerQA-Hard-035\newline
WebWalkerQA-Hard-036, WebWalkerQA-Hard-037, WebWalkerQA-Hard-038, WebWalkerQA-Hard-039, WebWalkerQA-Hard-040, WebWalkerQA-Hard-041, WebWalkerQA-Hard-042, WebWalkerQA-Hard-043, WebWalkerQA-Hard-044, WebWalkerQA-Hard-045, WebWalkerQA-Hard-046, WebWalkerQA-Hard-047, WebWalkerQA-Hard-048, WebWalkerQA-Hard-049, WebWalkerQA-Hard-050\newline
WebWalkerQA-Hard-051, WebWalkerQA-Hard-052, WebWalkerQA-Hard-053, WebWalkerQA-Hard-054, WebWalkerQA-Hard-055, WebWalkerQA-Hard-056, WebWalkerQA-Hard-057, WebWalkerQA-Hard-058, WebWalkerQA-Hard-059, WebWalkerQA-Hard-060, WebWalkerQA-Hard-061, WebWalkerQA-Hard-062, WebWalkerQA-Hard-063, WebWalkerQA-Hard-064, WebWalkerQA-Hard-065\newline
WebWalkerQA-Hard-066, WebWalkerQA-Hard-067, WebWalkerQA-Hard-068, WebWalkerQA-Hard-069, WebWalkerQA-Hard-070, WebWalkerQA-Hard-071, WebWalkerQA-Hard-073, WebWalkerQA-Hard-074, WebWalkerQA-Hard-075, WebWalkerQA-Hard-076 \\
\midrule
DeepSearchQA & Sequential &
DeepSearchQA-004, DeepSearchQA-083, DeepSearchQA-095, DeepSearchQA-131, DeepSearchQA-257, DeepSearchQA-268, DeepSearchQA-300, DeepSearchQA-316, DeepSearchQA-349, DeepSearchQA-370, DeepSearchQA-403, DeepSearchQA-450, DeepSearchQA-517, DeepSearchQA-521, DeepSearchQA-568\newline
DeepSearchQA-605, DeepSearchQA-609, DeepSearchQA-680, DeepSearchQA-719, DeepSearchQA-759, DeepSearchQA-763, DeepSearchQA-828, DeepSearchQA-848, DeepSearchQA-875, DeepSearchQA-885 \\
\addlinespace[2pt]
DeepSearchQA & Parallel &
DeepSearchQA-012, DeepSearchQA-053, DeepSearchQA-070, DeepSearchQA-106, DeepSearchQA-110, DeepSearchQA-193, DeepSearchQA-203, DeepSearchQA-379, DeepSearchQA-408, DeepSearchQA-460, DeepSearchQA-482, DeepSearchQA-487, DeepSearchQA-534, DeepSearchQA-561, DeepSearchQA-577\newline
DeepSearchQA-589, DeepSearchQA-614, DeepSearchQA-652, DeepSearchQA-666, DeepSearchQA-682, DeepSearchQA-703, DeepSearchQA-728, DeepSearchQA-876, DeepSearchQA-880, DeepSearchQA-888 \\
\bottomrule
\end{tabularx}
\caption{Question IDs used in the main experiments. BrowseComp IDs follow the official order after filtering geography-only direct-field questions; WebWalker-Hard-English IDs are the direct-anchor subset; DeepSearchQA IDs are sampled from the classifier-assigned sequential and parallel pools.}
\label{tab:benchmark-question-ids}
\end{table}

\subsection{DAG Construction and Audit Prompts}
\label{app:dag-prompts}

This appendix summarizes the LLM prompts used by the structured DAG construction pipeline. The production runner is \texttt{run\_query\_subgraph\_v4.}\allowbreak\texttt{20260314\_032109.py}. We show the module-level prompts rather than raw trajectory payloads, which vary by instance.

\begin{promptbox}{Runner Prompt: Q0 Unit Decomposition}
\textbf{Role.} Decompose a user question into atomic Q0 units for a search-trajectory DAG.

\textbf{Task.} Each unit is one testable constraint or requested field from the question. Unit types include target type, location, attribute, numeric constraint, answer field, and temporal constraint.

\textbf{Output.} Return JSON with a list of units, each containing \texttt{unit\_id}, \texttt{unit\_type}, and \texttt{q0\_span}.

\textbf{Rules.} Be exhaustive; capture all constraints and requested fields; keep units atomic; make \texttt{q0\_span} a verbatim or near-verbatim substring of the question; output valid JSON only.
\end{promptbox}

\begin{promptbox}{Runner Prompt: Q0-to-Query Match Arbiter}
\textbf{Role.} Judge whether a search query operationalizes, addresses, or investigates a specific constraint unit from Q0.

\textbf{Definition.} Operationalizes means the query is purposefully trying to find information related to that constraint, including through paraphrase, synonym, or a narrower/broader formulation.

\textbf{Reject.} Surface word overlap alone is not sufficient; queries that share common words without targeting the constraint are rejected.

\textbf{Output.} For each pair, return JSON with \texttt{pair\_id}, \texttt{match} as true/false, and a one-sentence reason. Output valid JSON only.
\end{promptbox}

\begin{promptbox}{Runner Prompt: Q0 Paraphrase / PK Attribution Rule}
\textbf{Role.} Judge whether a token or short phrase in a query is only a paraphrase of information already present in Q0.

\textbf{Labels.} Return either \texttt{q0\_paraphrase} or \texttt{truly\_new}.

\textbf{Rules.} Judge meaning rather than exact wording. Small wording changes, tense changes, inflections, and near-synonyms can still be Q0 paraphrases. Return \texttt{truly\_new} if the token adds a more specific fact, a different fact, or a new entity.
\end{promptbox}

\begin{promptbox}{Runner Prompt: Answer Unit Decomposition}
\textbf{Role.} Decompose the final answer into atomic answer units for a search-trajectory DAG.

\textbf{Task.} Each unit is one independently verifiable factual claim, such as an entity name, date, number, descriptive attribute, or short description. The original question is provided to identify which claims are relevant.

\textbf{Output.} Return JSON with \texttt{answer\_units}; each unit contains \texttt{unit\_id}, \texttt{claim}, and \texttt{unit\_type}.

\textbf{Rules.} Capture all distinct factual claims; keep each claim short and verifiable; output valid JSON only.
\end{promptbox}

\begin{promptbox}{Runner Prompt: Answer Support Match}
\textbf{Role.} Judge whether a query's retrieved evidence supports a normalized answer unit.

\textbf{Input.} The prompt receives the original question, the final answer's supporting analysis, and candidate pairs of answer unit and query evidence.

\textbf{Accept.} Mark \texttt{support=true} only when the evidence excerpt materially supports the answer unit, including via alias, paraphrase, or short cross-sentence inference.

\textbf{Reject.} Reject topical overlap, overlapping words without claim support, and claims asserted only in the answer reasoning rather than in query evidence.

\textbf{Output.} Return JSON verdicts with \texttt{pair\_id}, \texttt{supports}, \texttt{confidence}, \texttt{reason}, and the best short supporting evidence sentence.
\end{promptbox}

\begin{promptbox}{Runner Prompt: Query-to-Query / PK / Failure-Response Edge Attribution}
\textbf{Role.} Act as a Query-DAG edge classifier for one search turn. Output only edges whose target is a query issued in the current turn.

\textbf{Input.} The prompt receives Q0 as background, prior query index, turn-level status, current reasoning block split into sentence IDs, current-turn queries, and a hint chart containing token statuses (\texttt{Q0\_ONLY}, \texttt{Q0\_SEEN}, \texttt{SEEN}, \texttt{NEW}), used signals, reasoning sentence hits, and provenance windows from earlier snippets or visits.

\textbf{Allowed edges.} Evidence-use edges from prior query to current query are allowed only when a used signal has provenance attributed to that prior query or an explicit reasoning sentence ties the signal to it. Prior-knowledge edges are allowed only when a signal is new, appears in reasoning support, and has no provenance. Soft failure-response edges are allowed only when the reasoning explicitly states that a prior query was insufficient and the current query fills that gap.

\textbf{Hard gates.} Every emitted edge must cite a reasoning sentence ID or provenance window. Do not output Q0 edges, same-turn edges, deterministic hard failure-response edges, hallucinated sources, or PK fallback when provenance exists.

\textbf{Parent selection.} Use informative tokens only, ignore generic search-template tokens, prefer visits over snippets, and keep the minimal parent set whose union covers the informative used signals. Each retained parent must explain at least one signal not already covered by another parent.

\textbf{Output.} Return JSON with \texttt{edges} and \texttt{no\_source\_found}. Each edge includes source, target, edge kind, optional failure subtype, covered signals, evidence, and confidence.
\end{promptbox}

\begin{promptbox}{Constraint-Unit Annotation Rules for BCG and DCC}
\textbf{Goal.} Produce semantic constraint units used by the constraint-grounding metrics, then determine which query nodes deploy each unit.

\textbf{Unit creation.} We first decompose the question into semantic units such as entity, attribute, time, role, location, source, ranking, threshold, or requested answer field. The decomposition is LLM-assisted and manually reviewed.

\textbf{Query deployment.} A query is marked as deploying a unit by a rule-based lexical overlap test after basic normalization (lowercasing, singularization, stopword/generic-term removal). Generic-word overlaps are ignored, and for multi-token units numeric-only or weak single-token overlaps do not count without distinctive non-numeric support. 

\textbf{Metric use.} For sequential questions, BCG (backbone constraint share) counts whether these unit deployments fall on the answer backbone rather than side branches. For parallel questions, DCC (direct constraint coverage) counts how many distinct question units are covered by direct answer-supporting queries.
\end{promptbox}

\begin{promptbox}{Outlier Mechanism Judge Prompt (\S\ref{subsec:metric-boundary})}
\textbf{Role.} Label mechanisms for structural metric outliers in search-agent DAG analysis.

\textbf{Input.} The judge receives the outlier type, benchmark, regime, model, case ID, correctness, metric values, question, gold answer, model prediction, final answer from messages, DAG summary, transcript excerpt, and, when available, the full or pruned raw trajectory and raw DAG item.

\textbf{Labels.} Use exactly one main label: \texttt{WRONG\_TARGET\_BINDING}, \texttt{SEMANTIC\_SPECIFICITY\_FAILURE}, \texttt{SHORTCUT\_SUCCESS}, \texttt{OVER\_SEARCH\_SUCCESS}, or \texttt{UNCERTAIN}.

\textbf{Outlier constraints.} High-score wrong cases can only be wrong-target binding, semantic-specificity failure, or uncertain. Low-score correct cases can only be shortcut success, over-search success, or uncertain.

\textbf{Rules.} Base the label on the trajectory/messages and DAG first. Use metric values only for orientation. For shortcut success, also mark \texttt{internal\_knowledge}, \texttt{open\_web\_shortcut}, \texttt{mixed}, or \texttt{unclear}.

\textbf{Output.} Return one JSON object with \texttt{mechanism\_main}, \texttt{shortcut\_subtype}, \texttt{confidence}, \texttt{needs\_human\_check}, \texttt{rationale}, \texttt{evidence\_basis}, and \texttt{ambiguity\_notes}. Set \texttt{needs\_human\_check=true} if confidence is below 0.70, if transcript/DAG context is missing, or if two mechanisms are similarly plausible.
\end{promptbox}

\FloatBarrier
\subsection{Structural Profiles Across Question Types and Agents}
\label{subsec:structural-profiles}
\begin{figure}[htbp]
  \centering
  \IfFileExists{figures/depth-width_plot.pdf}{%
    \includegraphics[width=\linewidth]{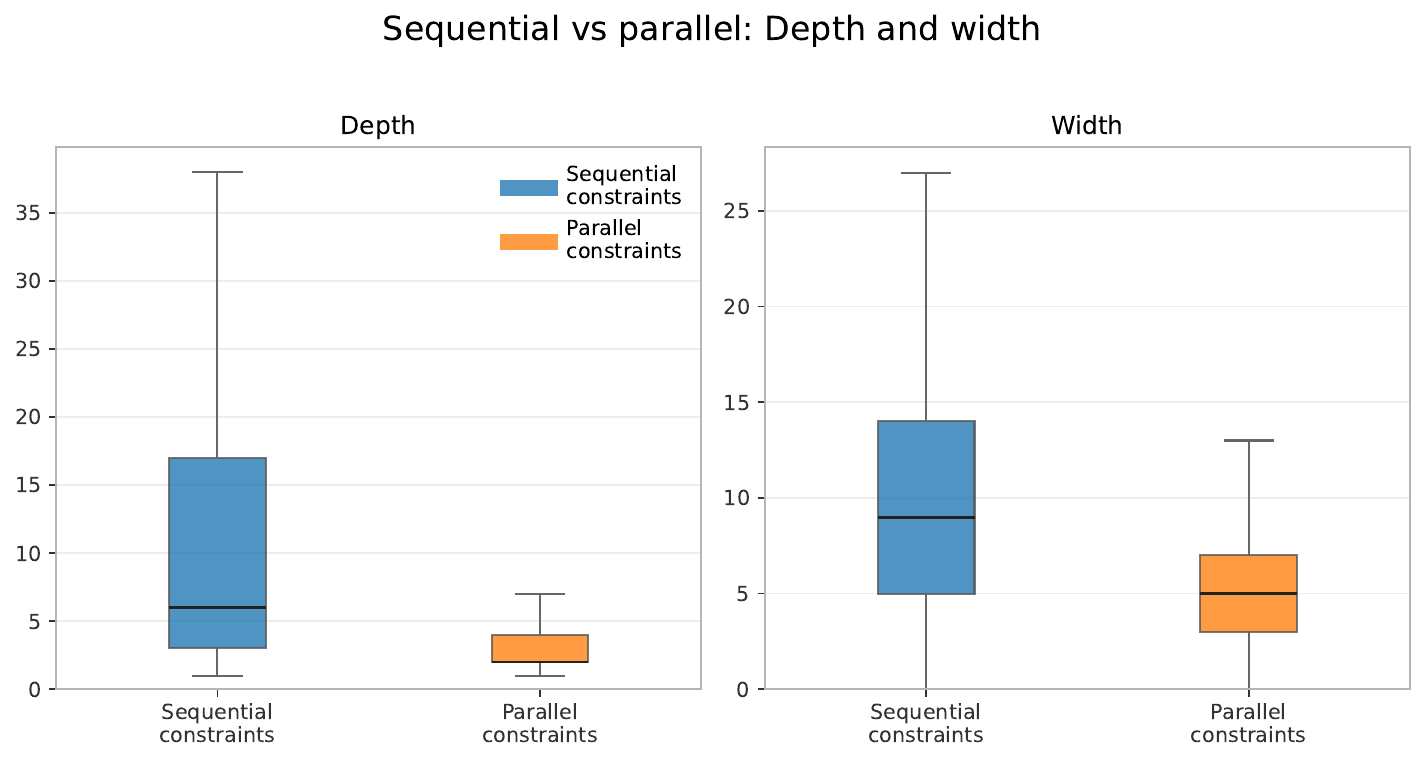}
  }{%
    \fbox{\parbox{0.92\linewidth}{\centering Placeholder for structural profile boxplots.}}
  }
  \caption{Boxplots of depth and width metrics for sequential and parallel constraint questions.}
  \label{fig:question-profile-other-metrics}
\end{figure}

\begin{figure}[htbp]
  \centering
  \IfFileExists{figures/other_three_metrics_plot.pdf}{%
    \includegraphics[width=\linewidth]{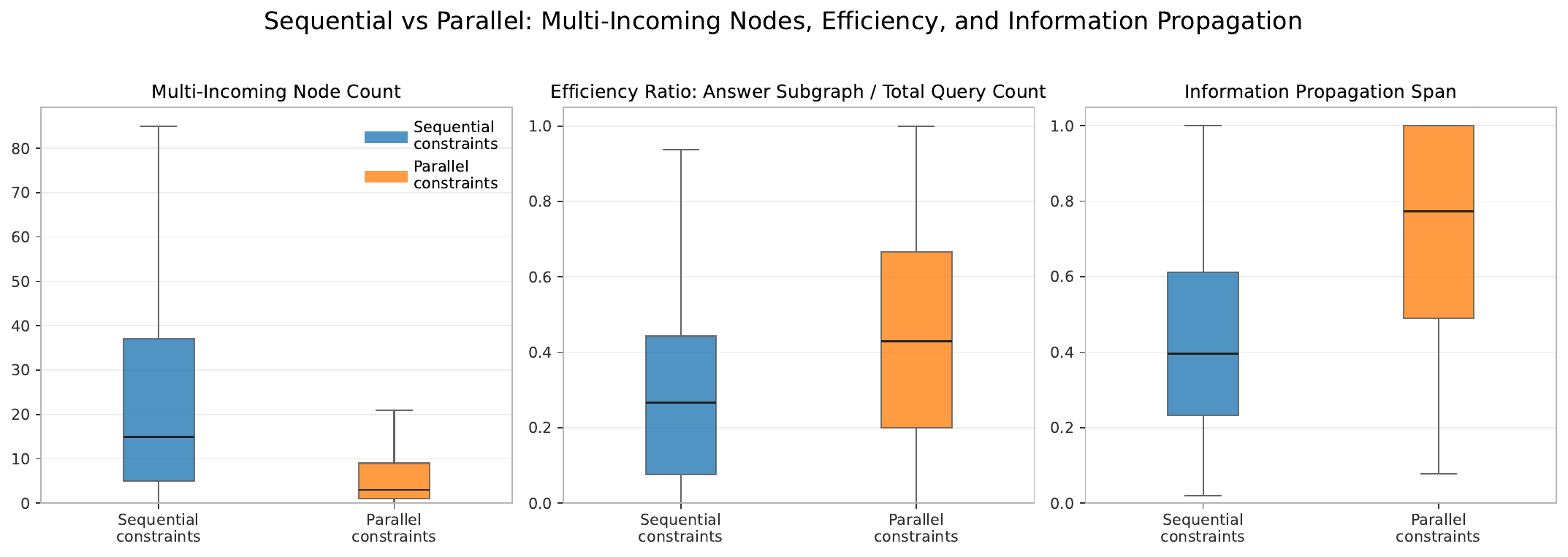}
  }{%
    \fbox{\parbox{0.92\linewidth}{\centering Placeholder for structural profile boxplots.}}
  }
  \caption{Boxplots of three trajectory-level metrics for sequential and parallel constraint questions: multi-incoming node count, efficiency ratio, and evidence-reuse span.}
  \label{fig:question-profile-metrics}
\end{figure}

\paragraph{Sequential constraints deepen the DAG and increase branching.}
Sequential-constraint questions require resolving intermediate entities before later constraints become interpretable. This tends to produce longer evidence chains and more branching than parallel-constraint questions, where several constraints can often be checked against a common candidate or source. The following profile metrics show where this additional structure appears.

\paragraph{Sequential questions reduce query efficiency.}
\emph{Query efficiency} measures what fraction of the agent's search queries
ultimately contribute to the answer, as opposed to being spent on searching
candidates that are later discarded, measured by the ratio of queries that
lie on a path leading to the final answer to the total number of
queries issued.
Across all models, efficiency drops noticeably when questions involve sequential constraints that require resolving more latent entities and verifying more constraints.
The remaining queries are not wasted at random: they concentrate on
two recognizable activities, generating candidate entities by trial
and error, and subsequent verification of candidates that turn out not
to satisfy every constraint.

\paragraph{Sequential questions require more multi-source synthesis.}
The set of incoming edges to a query node tells us how that query was
formulated: a node with a single parent inherits its content from a
single prior query's retrieval result, while a node with multiple parents synthesizes information gathered across several earlier queries. We therefore count, for each trajectory, the number of query nodes with two or more incoming edges---queries that aggregate evidence rather than extend a single line of inquiry. On sequential-constraint questions, the agent increasingly issues queries that fuse multiple constraints or cross-check candidate entities against several pieces of prior evidence at once, rather than pursuing one constraint at a time.

\paragraph{Evidence is reused farther in parallel-constraint questions.}
In parallel-constraint trajectories, evidence retrieved by
a query is attributed to later queries throughout a much
larger fraction of the remaining search; in
sequential-constraint trajectories, retrieved evidence is
usually attributed only to the next few turns before the agent
moves on to the next sub-problem. We quantify this with the
\emph{normalized evidence-reuse span} of each parent--child edge
$u \rightarrow v$:
\[
\frac{\text{turn}(v) - \text{turn}(u)}{\text{total\_turns} -
\text{turn}(u)},
\]
the gap in search turns expressed as a fraction of the
trajectory still available after $u$ was issued. The parallel
group has a much higher mean and median than the sequential group. 

\FloatBarrier

\subsection{Accuracy-Metric Ablations}
\label{app:metric-ablation}

This appendix reports the quantitative checks behind the three-module process analysis in Section~\ref{subsec:accuracy-metrics}. All metrics are oriented so that larger values should rank correct trajectories above incorrect trajectories: topology and grounding enter positively, while prior-knowledge risk enters with the negative sign used by the fixed additive score. BrowseComp is evaluated as the sequential regime, WebWalker-Hard as the parallel regime, and DeepSearchQA is split into sequential and parallel subsets.

\begin{table}[htbp]
\centering
\small
\setlength{\tabcolsep}{3pt}
\begin{tabular}{llccc}
\toprule
Dataset / regime & Module & Topology & Grounding & PK-risk signal \\
\midrule
BrowseComp / sequential & AUC & 0.714 & 0.764 & 0.707 \\
                         & AP  & 0.530 & 0.503 & 0.405 \\
WebWalker-Hard / parallel     & AUC & 0.746 & 0.680 & 0.634 \\
                         & AP  & 0.848 & 0.827 & 0.798 \\
DeepSearchQA / sequential & AUC & 0.734 & 0.808 & 0.490 \\
                          & AP  & 0.713 & 0.773 & 0.448 \\
DeepSearchQA / parallel  & AUC & 0.771 & 0.866 & 0.305 \\
                         & AP  & 0.601 & 0.760 & 0.362 \\
\bottomrule
\end{tabular}
\caption{Single-module diagnostic results, macro-averaged across models. The PK column reports the signed PK-risk signal used by the additive score, not the raw amount of PK. The table shows that individual modules are associated with answer outcomes, but also that no single module is uniformly sufficient across regimes.}
\label{tab:single-module-diagnostics}
\end{table}

\begin{table}[htbp]
\centering
\small
\setlength{\tabcolsep}{3pt}
\begin{tabular}{lccc}
\toprule
Dataset / regime & Topology only & + Grounding & + PK-risk penalty \\
\midrule
BrowseComp / sequential      & 0.714 / 0.530 & 0.816 / 0.574 & 0.855 / 0.617 \\
WebWalker-Hard / parallel         & 0.746 / 0.848 & 0.779 / 0.881 & 0.840 / 0.939 \\
DeepSearchQA / sequential    & 0.734 / 0.713 & 0.840 / 0.774 & 0.844 / 0.809 \\
DeepSearchQA / parallel      & 0.771 / 0.601 & 0.830 / 0.691 & 0.856 / 0.726 \\
\bottomrule
\end{tabular}
\caption{Stage-wise additive-score ablation. Each cell reports AUC / AP, macro-averaged across models. The main-text table reports AUC and incremental gains; this table adds the AP values.}
\label{tab:stagewise-ablation-auc-ap}
\end{table}

\begin{table}[htbp]
\centering
\small
\setlength{\tabcolsep}{4pt}
\begin{tabular}{llrrrr}
\toprule
Dataset / regime & Model & Acc. & Topology & + Grounding & + PK-risk penalty \\
\midrule
BrowseComp / sequential & TYDP & 0.493 & 0.742 & 0.868 & 0.898 \\
BrowseComp / sequential & MiroThinker & 0.393 & 0.884 & 0.888 & 0.897 \\
BrowseComp / sequential & WebSailor & 0.140 & 0.787 & 0.825 & 0.878 \\
BrowseComp / sequential & TYDP-GPT5 & 0.453 & 0.528 & 0.646 & 0.720 \\
BrowseComp / sequential & TYDP-Qwen3 & 0.040 & 0.627 & 0.852 & 0.880 \\
\midrule
WebWalker-Hard / parallel & TYDP & 0.843 & 0.714 & 0.741 & 0.822 \\
WebWalker-Hard / parallel & MiroThinker & 0.786 & 0.728 & 0.747 & 0.774 \\
WebWalker-Hard / parallel & WebSailor & 0.686 & 0.836 & 0.857 & 0.884 \\
WebWalker-Hard / parallel & TYDP-GPT5 & 0.814 & 0.704 & 0.794 & 0.826 \\
WebWalker-Hard / parallel & TYDP-Qwen3 & 0.443 & 0.749 & 0.758 & 0.895 \\
\midrule
DeepSearchQA / sequential & TYDP & 0.520 & 0.744 & 0.827 & 0.827 \\
DeepSearchQA / sequential & MiroThinker & 0.440 & 0.711 & 0.724 & 0.724 \\
DeepSearchQA / sequential & WebSailor & 0.200 & 0.765 & 0.840 & 0.860 \\
DeepSearchQA / sequential & TYDP-GPT5 & 0.640 & 0.736 & 0.986 & 0.986 \\
DeepSearchQA / sequential & TYDP-Qwen3 & 0.320 & 0.713 & 0.824 & 0.824 \\
\midrule
DeepSearchQA / parallel & TYDP & 0.360 & 0.733 & 0.816 & 0.858 \\
DeepSearchQA / parallel & MiroThinker & 0.400 & 0.623 & 0.730 & 0.763 \\
DeepSearchQA / parallel & WebSailor & 0.160 & 0.810 & 0.940 & 0.964 \\
DeepSearchQA / parallel & TYDP-GPT5 & 0.640 & 0.802 & 0.802 & 0.802 \\
DeepSearchQA / parallel & TYDP-Qwen3 & 0.120 & 0.886 & 0.864 & 0.894 \\
\bottomrule
\end{tabular}
\caption{Per-model numerical values underlying \autoref{tab:stagewise-auc-main}. Every AUC is computed within one fixed model and benchmark-regime, comparing only that model's correct and incorrect trajectories; no cross-model trajectory pair enters an individual AUC. The columns report cumulative held-out AUC after adding the three modules. For sequential regimes, the modules are answer-backbone concentration, backbone constraint share, and sequential PK penalty; for parallel regimes, they are answer-path directness, direct constraint coverage, and parallel PK score. Acc. is the empirical answer accuracy for that benchmark-regime and model.}
\label{tab:stagewise-main-per-model}
\end{table}

\subsection{Score Calibration and Auxiliary Outcome Prediction}
\label{app:score-calibration}

AUC evaluates the association between the behavioral diagnostics and answer outcomes at the ranking level. As an auxiliary outcome-risk check, we also ask whether the fixed DAG-derived score produces interpretable score bands and whether a held-out threshold preserves this association at the trajectory level. We compute composite DAG-derived scores out-of-fold under the same question-held-out protocol as the main result. For score bands, we sort trajectories by the composite score within each benchmark-regime and report empirical accuracy in score tertiles. For threshold prediction, each benchmark-regime/model learns a threshold on training questions that maximizes balanced accuracy, then applies it to held-out questions.

\begin{table}[htbp]
\centering
\small
\setlength{\tabcolsep}{4pt}
\begin{tabular}{lccc}
\toprule
Dataset / regime & Bottom 33\% & Middle 33\% & Top 33\% \\
\midrule
BrowseComp / sequential        & 8.4 & 27.2 & 55.6 \\
WebWalker-Hard / parallel      & 47.0 & 72.6 & 94.8 \\
DeepSearchQA / sequential      & 16.7 & 38.1 & 73.2 \\
DeepSearchQA / parallel        & 4.8  & 33.3 & 63.4 \\
\bottomrule
\end{tabular}
\caption{Answer accuracy by composite DAG-derived score tertile within each benchmark-regime. Values are percentages.}
\label{tab:score-band-tertiles}
\end{table}

\begin{table}[htbp]
\centering
\small
\setlength{\tabcolsep}{4pt}
\begin{tabular}{lrrrrrr}
\toprule
Model & Seq. bottom 20\% & Seq. top 20\% & Seq. lift & Par. bottom 20\% & Par. top 20\% & Par. lift \\
\midrule
TYDP      & 8.6 & 94.3 & +85.7 & 57.9 & 94.7 & +36.8 \\
MiroThinker & 5.7  & 94.3 & +88.6 & 52.6 & 94.7 & +42.1 \\
WebSailor & 0.0  & 48.6 & +48.6 & 15.8 & 94.7 & +78.9 \\
TYDP-GPT5  & 8.6 & 65.7 & +57.1 & 47.4 & 100.0 & +52.6 \\
TYDP-Qwen3 & 0.0  & 20.0 & +20.0 & 5.3  & 78.9 & +73.7 \\
\bottomrule
\end{tabular}
\caption{Top- and bottom-quintile accuracy by model and regime. Values are percentages. The score separates high-risk and low-risk trajectories for every model, but the absolute accuracy of high-score sequential trajectories remains model-dependent.}
\label{tab:score-band-model-quintiles}
\end{table}

\begin{table}[htbp]
\centering
\small
\setlength{\tabcolsep}{3pt}
\begin{tabular}{lrrrrrr}
\toprule
Dataset / regime & Prev. & Acc. & Bal. Acc. & Prec. & Recall & $F_1$ \\
\midrule
BrowseComp / sequential        & 30.4 & 80.8 & 80.2 & 65.3 & 78.5 & 71.3 \\
WebWalker-Hard / parallel      & 71.4 & 72.9 & 73.2 & 87.4 & 72.4 & 79.2 \\
DeepSearchQA / sequential      & 42.4 & 68.8 & 69.7 & 60.6 & 75.5 & 67.2 \\
DeepSearchQA / parallel        & 33.6 & 80.8 & 78.5 & 71.4 & 71.4 & 71.4 \\
\bottomrule
\end{tabular}
\caption{Auxiliary held-out outcome-risk prediction from the fixed DAG-derived score within each benchmark-regime. Values are percentages. Thresholds are selected only on training questions within each benchmark-regime/model and then evaluated on held-out questions. Aggregate $F_1$ is $70.5$ for sequential and $78.0$ for parallel questions.}
\label{tab:score-threshold-prediction}
\end{table}

These auxiliary results clarify how the behavioral score relates to outcomes. High-score bands are much more accurate than low-score bands, so the score can support trajectory-level risk assessment and selective review without serving as a standalone correctness judgment. At the same time, sequential precision is lower because a coherent answer-support path can still bind to the wrong target; this behavior--outcome divergence is analyzed in Section~\ref{subsec:metric-boundary}.

\subsection{Non-DAG LLM Judge Baselines}
\label{app:llm-judge-baseline}

As an auxiliary outcome-risk comparison, we compare the thresholded DAG-derived score with two non-DAG LLM judges. The first is an outcome-oriented baseline that asks GPT-5.2 to judge whether the trajectory's final answer is correct from the full trajectory. This tests whether the DAG adds value beyond having a strong model summarize the entire log. The second is an ordered-query-list baseline that gives the judge only the original question, final answer, and chronological query list, with retrieved evidence, page visits, reasoning, and DAG edges removed. This tests whether query order alone is enough without explicit dependency structure. Table~\ref{tab:llm-judge-summary} gives the pooled comparison. Macro-$F_1$ weights correct and incorrect trajectories equally, while positive $F_1$ treats correct trajectories as the positive class. \textsc{SearchAtlas}improves over the full-trajectory judge by $3.8$ accuracy and $6.7$ macro-$F_1$ points, and over the ordered-query-list judge by $9.9$ and $16.4$ points. To make the more detailed comparison with thresholded DAG-derived predictions direct, AUC and AP in Table~\ref{tab:llm-judge-baseline} are computed from the binary predictions only.

\begin{table}[htbp]
\centering
\small
\setlength{\tabcolsep}{10pt}
\begin{tabular}{lrrr}
\toprule
Method & Accuracy & Macro-$F_1$ & Positive $F_1$ \\
\midrule
Ordered-query-list judge & 67.7 & 60.8 & 44.4 \\
Full-trajectory judge & 73.8 & 70.5 & 60.6 \\
SearchAtlas & 77.6 & 77.2 & 74.0 \\
\midrule
$\Delta$ over ordered list & +9.9 & +16.4 & +29.6 \\
$\Delta$ over full trajectory & +3.8 & +6.7 & +13.4 \\
\bottomrule
\end{tabular}
\caption{Pooled trajectory-level comparison on all 1,350 trajectories. Values are percentages.}
\label{tab:llm-judge-summary}
\end{table}

\begin{promptbox}{Full-Trajectory GPT-5.2 Correctness Judge Prompt}
\textbf{Role.} You are a trajectory-structure correctness judge for a web-search research agent.

\textbf{Task.} Predict whether the agent's final answer is correct using only the information inside the provided trajectory and the structure of the agent's search/reasoning process.

\textbf{Hard restrictions.} Do not use your own world knowledge, memory, or web knowledge. Do not solve the original question yourself. Treat all facts as unknown unless supported by the trajectory text. The gold answer is not provided.

\textbf{Allowed signals.} Use structural evidence such as whether constraints were searched, whether evidence was found, whether the final answer is grounded in cited/tool evidence, unresolved contradictions, hallucinated leaps, failed searches, or unsupported candidate switches. Do not reward verbosity.

\textbf{Binary decision.} Choose exactly one final label: \texttt{correct} or \texttt{incorrect}. Do not output \texttt{uncertain}. If evidence is incomplete or ambiguous, still make the best binary prediction from trajectory structure alone and use confidence to express uncertainty.

\textbf{Output.} Return JSON only with fields: \texttt{prediction}, \texttt{confidence}, \texttt{final\_answer\_supported}, \texttt{all\_question\_constraints\_checked}, \texttt{uses\_only\_trajectory\_evidence}, \texttt{gold\_answer\_present\_in\_prompt}, \texttt{structural\_signals}, and \texttt{rationale}.

\textbf{User payload.} Task ID; input file; agent; original question; agent final answer to judge; search transcript, introduced as ``the only evidence you may use''; return JSON only.
\end{promptbox}

\begin{table}[htbp]
\centering
\small
\setlength{\tabcolsep}{4pt}
\begin{tabular}{llrrrrrrr}
\toprule
Scope & Method & $N$ & Acc. & Prec. & Recall & Pos. $F_1$ & AUC & AP \\
\midrule
All        & Thresholded DAG-derived score (SearchAtlas)   & 1350 & 77.6 & 73.0 & 75.0 & 74.0 & 77.3 & 65.4 \\
All        & Full-trajectory LLM judge                     & 1350 & 73.8 & 83.7 & 47.5 & 60.6 & 70.3 & 62.0 \\
All        & Ordered-query-list LLM judge                  & 1350 & 67.7 & 82.5 & 30.4 & 44.4 & 62.8 & 54.6 \\
\midrule
Sequential & Thresholded DAG-derived score (SearchAtlas)   &  875 & 79.1 & 64.4 & 77.9 & \textbf{70.5} & \textbf{78.8} & 57.3 \\
Sequential & Full-trajectory LLM judge                     &  875 & 76.8 & 82.0 & 35.6 & 49.6 & 65.9 & 49.9 \\
Sequential & Ordered-query-list LLM judge                  &  875 & 75.3 & 82.8 & 29.2 & 43.2 & 63.2 & 46.9 \\
\midrule
Parallel   & Thresholded DAG-derived score (SearchAtlas)   &  475 & 74.9 & 84.7 & 72.3 & \textbf{78.0} & 75.7 & \textbf{78.3} \\
Parallel   & Full-trajectory LLM judge                     &  475 & 68.2 & 84.7 & 58.9 & 69.5 & 71.0 & 75.2 \\
Parallel   & Ordered-query-list LLM judge                  &  475 & 53.7 & 82.1 & 31.5 & 45.5 & 60.3 & 68.0 \\
\bottomrule
\end{tabular}
\caption{Trajectory-level comparison between the thresholded DAG-derived score
and non-DAG LLM-judge baselines on all three benchmarks (BrowseComp 150 questions
sequential, WebWalker-Hard 70 questions parallel, DeepSearchQA 25
sequential and 25 parallel questions; all crossed with five agents). Values are percentages except $N$. Precision, recall, and $F_1$ treat correct trajectories as the positive
class. AUC and AP in this table use binary predictions so that all rows share
the same thresholded-decision protocol.}
\label{tab:llm-judge-baseline}
\end{table}

\begin{promptbox}{Ordered-Query-List GPT-5.2 Judge Prompt}
\textbf{Role.} You are a trajectory-structure correctness judge for a web-search research agent, but you may only inspect the ordered query list.

\textbf{Task.} Predict whether the agent's final answer is correct using the original question, the agent's final answer, and the chronological list of issued search queries.

\textbf{Unavailable information.} Retrieved search results, visited pages, reasoning text, citations, and DAG edges are deliberately removed. Do not infer hidden evidence from query wording alone.

\textbf{Allowed signals.} Use only coarse query-list signals such as whether the agent searched for the stated constraints, whether the query sequence appears to refine or drift from the task, whether key entities in the final answer appear in the query list, and whether the query list looks sufficient or under-specified.

\textbf{Binary decision.} Choose exactly one final label: \texttt{correct} or \texttt{incorrect}. If the query list is insufficient to tell, still make the best binary prediction and use confidence to express uncertainty.

\textbf{Output.} Return JSON only with \texttt{prediction}, \texttt{confidence}, \texttt{query\_coverage}, \texttt{final\_answer\_appears\_searched}, \texttt{query\_sequence\_sufficiency}, \texttt{structural\_signals}, and \texttt{rationale}.
\end{promptbox}

Table~\ref{tab:ordered-query-list-full} reports the full ordered-query-list results from the confidence-ranked judge score. The baseline is conservative: it attains high precision because it predicts relatively few trajectories as correct, but its recall is low across both sequential and parallel regimes. Thus, chronological query text can identify a small set of obvious successes, but without dependency edges it misses many trajectories whose final answers are supported by retrieved evidence.

\begin{table}[htbp]
\centering
\small
\setlength{\tabcolsep}{4pt}
\begin{tabular}{lrrrrrrr}
\toprule
Dataset / regime & $N$ & Acc. & Prec. & Recall & $F_1$ & AUC & AP \\
\midrule
BrowseComp / sequential        & 750 & 77.9 & 69.5 & 28.3 & 40.1 & 77.1 & 56.1 \\
DeepSearchQA / parallel        & 125 & 69.6 & 70.0 & 12.7 & 19.9 & 69.0 & 50.7 \\
DeepSearchQA / sequential      & 125 & 60.0 & 50.0 & 14.7 & 19.8 & 66.3 & 58.3 \\
WebWalker-Hard / parallel      & 350 & 48.0 & 84.8 & 32.5 & 44.1 & 60.7 & 78.9 \\
\midrule
All                             & 1350 & 67.7 & 82.5 & 30.4 & 44.4 & 79.4 & 70.2 \\
\bottomrule
\end{tabular}
\caption{Ordered-query-list judge results. The benchmark-regime rows are per-model macro averages within each benchmark-regime. The All row is pooled over all 1,350 trajectories. AUC and AP in this table use the judge's confidence-oriented score, unlike the binary-prediction AUC/AP in Table~\ref{tab:llm-judge-baseline}. Values are percentages except $N$.}
\label{tab:ordered-query-list-full}
\end{table}

\begin{table}[htbp]
\centering
\small
\setlength{\tabcolsep}{4pt}
\begin{tabular}{llrrrr}
\toprule
Dataset / regime & Model & $N$ & Acc. & $F_1$ & AUC / AP \\
\midrule
BrowseComp / sequential & TYDP-GPT5 & 150 & 66.0 & 47.4 & 65.1 / 65.2 \\
BrowseComp / sequential & MiroThinker & 150 & 76.7 & 58.8 & 80.1 / 75.1 \\
BrowseComp / sequential & TYDP-Qwen3 & 150 & 94.7 & 0.0 & 78.1 / 10.4 \\
BrowseComp / sequential & TYDP & 150 & 61.3 & 40.8 & 81.8 / 76.3 \\
BrowseComp / sequential & WebSailor & 150 & 90.7 & 53.3 & 80.3 / 53.4 \\
\midrule
DeepSearchQA / parallel & TYDP-GPT5 & 25 & 40.0 & 11.8 & 58.0 / 70.9 \\
DeepSearchQA / parallel & MiroThinker & 25 & 64.0 & 18.2 & 62.7 / 52.4 \\
DeepSearchQA / parallel & TYDP-Qwen3 & 25 & 88.0 & 0.0 & 78.0 / 26.5 \\
DeepSearchQA / parallel & TYDP & 25 & 72.0 & 36.4 & 67.7 / 58.6 \\
DeepSearchQA / parallel & WebSailor & 25 & 84.0 & 33.3 & 78.6 / 45.0 \\
\midrule
DeepSearchQA / sequential & TYDP-GPT5 & 25 & 40.0 & 11.8 & 72.9 / 78.6 \\
DeepSearchQA / sequential & MiroThinker & 25 & 68.0 & 42.9 & 51.0 / 58.1 \\
DeepSearchQA / sequential & TYDP-Qwen3 & 25 & 68.0 & 0.0 & 55.1 / 42.6 \\
DeepSearchQA / sequential & TYDP & 25 & 44.0 & 0.0 & 77.9 / 75.7 \\
DeepSearchQA / sequential & WebSailor & 25 & 80.0 & 44.4 & 74.5 / 36.7 \\
\midrule
WebWalker-Hard / parallel & TYDP-GPT5 & 70 & 55.7 & 68.7 & 47.4 / 80.5 \\
WebWalker-Hard / parallel & MiroThinker & 70 & 50.0 & 55.7 & 61.5 / 86.0 \\
WebWalker-Hard / parallel & TYDP-Qwen3 & 70 & 60.0 & 17.6 & 77.4 / 66.7 \\
WebWalker-Hard / parallel & TYDP & 70 & 28.6 & 34.2 & 55.2 / 86.9 \\
WebWalker-Hard / parallel & WebSailor & 70 & 45.7 & 44.1 & 62.0 / 74.5 \\
\bottomrule
\end{tabular}
\caption{Per-model ordered-query-list judge results. AUC and AP use the judge's confidence-oriented score. Values are percentages except $N$.}
\label{tab:ordered-query-list-per-model}
\end{table}
\paragraph{Trajectory-wide versus DAG-localized constraint grounding.}
\label{app:constraint-localization-baseline}
A trajectory-wide coverage baseline records the fraction of semantic question constraints deployed by at least one query anywhere in the run. It does not use dependency edges or distinguish answer-supporting queries from side branches. We compare it with the regime-specific DAG-derived grounding diagnostic computed from the same constraint units: backbone constraint share for sequential questions and direct constraint coverage at the answer-support frontier for parallel questions.

\begin{table}[htbp]
\centering
\small
\setlength{\tabcolsep}{7pt}
\begin{tabular}{lrrcc}
\toprule
Dataset / regime & $N$ & Agents & \shortstack{Trajectory-wide\\coverage} & \shortstack{DAG-localized\\grounding} \\
\midrule
BrowseComp / sequential   & 750 & 5 & 0.459 / 0.305 & 0.764 / 0.503 \\
WebWalker-Hard / parallel & 350 & 5 & 0.568 / 0.771 & 0.680 / 0.827 \\
\bottomrule
\end{tabular}
\caption{Constraint-grounding localization ablation. Each score is AUC / AP, macro-averaged across agents under question-held-out evaluation. The trajectory-wide baseline counts constraint coverage anywhere in the query sequence; DAG-localized grounding restricts the signal to the answer backbone (sequential) or answer-support frontier (parallel).}
\label{tab:constraint-localization-baseline}
\end{table}

Localizing constraint deployment improves both AUC and AP in both regimes. The larger BrowseComp difference also shows why repeated or scattered mentions of question constraints should not receive the same credit as constraints integrated into the answer-reaching evidence path.

\paragraph{Trajectory-wide versus answer-localized prior-knowledge reliance.}
\label{app:pk-localization-baseline}
We construct a matched trajectory-wide PK baseline from the same generated DAGs and PK attribution decisions. It counts all $\mathrm{PK}\to q$ edges in the trajectory, adds an indicator for an eligible $\mathrm{PK}\to A$ edge, and divides by the total number of queries. Unlike the current parallel PK score, it does not use answer reachability. We then replace only the PK term in the full additive score: topology, grounding, trajectories, agent-wise evaluation, semantic signs, and question-held-out folds remain fixed.

\begin{table}[htbp]
\centering
\small
\setlength{\tabcolsep}{5pt}
\begin{tabular}{lrccr}
\toprule
Dataset / regime & $N$ & \shortstack{Trajectory-wide\\PK} & \shortstack{Answer-specific /\\localized PK} & $\Delta$ AUC \\
\midrule
BrowseComp / sequential      & 750 & 0.735 / 0.511 & 0.855 / 0.617 & +0.119 \\
WebWalker-Hard / parallel    & 350 & 0.777 / 0.908 & 0.840 / 0.939 & +0.063 \\
DeepSearchQA / sequential    & 125 & 0.766 / 0.710 & 0.844 / 0.809 & +0.078 \\
DeepSearchQA / parallel      & 125 & 0.769 / 0.613 & 0.856 / 0.726 & +0.088 \\
\bottomrule
\end{tabular}
\caption{Matched PK-localization ablation. Each score is AUC / AP, macro-averaged across agents. Both columns retain identical topology and grounding terms; only the PK term is replaced. Positive $\Delta$ AUC favors the current answer-specific/localized PK term; deltas are computed from unrounded values.}
\label{tab:pk-localization-baseline}
\end{table}

The complete score is stronger with the answer-specific/localized PK term in all four benchmark-regimes. The parallel rows provide the direct DAG-localization test because the current parallel score restricts PK to the answer-supporting subgraph. In the sequential rows, the current term is an answer-specific pre-answer-context fallback check, so those rows establish the value of answer specificity but should not be interpreted as a graph-traversal ablation. PK-only performance is not uniformly stronger under localization; the result therefore supports localized PK as a complementary term alongside topology and grounding, rather than as a standalone correctness score.

\begin{table}[htbp]
\centering
\small
\setlength{\tabcolsep}{3pt}
\begin{tabular}{lcccc}
\toprule
Dataset / regime & + PK-risk penalty & w/o topology & w/o grounding & w/o PK \\
\midrule
BrowseComp / sequential      & 0.855 / 0.617 & 0.833 / 0.555 & 0.820 / 0.589 & 0.816 / 0.574 \\
WebWalker-Hard / parallel         & 0.840 / 0.939 & 0.749 / 0.900 & 0.806 / 0.895 & 0.779 / 0.881 \\
DeepSearchQA / sequential    & 0.844 / 0.809 & 0.795 / 0.732 & 0.726 / 0.693 & 0.840 / 0.774 \\
DeepSearchQA / parallel      & 0.856 / 0.726 & 0.710 / 0.682 & 0.517 / 0.495 & 0.830 / 0.691 \\
\bottomrule
\end{tabular}
\caption{Drop-one module ablations. Each cell reports AUC / AP. Removing any module usually reduces the complete three-module score, supporting the complementary role of topology, constraint grounding, and prior-knowledge reliance.}
\label{tab:dropone-ablation-auc-ap}
\end{table}

The ablations support two conclusions. First, topology is a useful first-order signal, but it is not a complete account of answer-support behavior. Constraint grounding and PK reliance surface outcome-associated distinctions that topology alone cannot see. Second, the contribution of each auxiliary module is regime-dependent: grounding is especially strong on sequential DeepSearchQA, while PK risk provides a large gain for the parallel WebWalker and DeepSearchQA subsets. Within the DeepSearchQA parallel split, TYDP-Qwen3 is the one model for which adding grounding slightly hurts AUC before the PK module is added; this is consistent with the role of grounding as a complementary rather than standalone signal.

\subsection{Raw-Log Process Baselines}
\label{app:raw-log-baseline}

This appendix tests whether the DAG-derived diagnostics are merely proxies for coarse trajectory effort, such as how many queries an agent issued, how many pages it visited, how long the trajectory was, or how often it repeated searches. These raw-log features carry modest accuracy-associated signal, confirming that search effort is not irrelevant. However, they conflate productive search with unproductive drift: the same increase in length or breadth can reflect either necessary evidence gathering or failed exploration. The DAG-derived diagnostics consistently improve over them because they evaluate the organization of evidence rather than the amount of search.

\begin{table}[htbp]
\centering
\small
\setlength{\tabcolsep}{5pt}
\begin{tabularx}{\textwidth}{@{}p{0.23\textwidth}p{0.22\textwidth}Y@{}}
\toprule
Metric & Meaning & Computation \\
\midrule
\texttt{log\_search\_queries} & number of search queries & count query strings in search tool calls, then \texttt{log1p} \\
\texttt{log\_visited\_pages} & number of visited pages & count URLs in visit tool calls, then \texttt{log1p} \\
\texttt{log\_trace\_length} & trajectory length & total non-system message characters, then \texttt{log1p} \\
\texttt{search\_redundancy} & repeated search-query ratio & $1-\frac{\text{unique normalized search queries}}{\text{search query count}}$ \\
\bottomrule
\end{tabularx}
\caption{Raw-log baseline features. These features are computed directly from the trajectory log and do not use the recovered DAG.}
\label{tab:raw-log-feature-defs}
\end{table}

We evaluate the log-only baselines with the same grouped 5-fold cross-validation protocol by question ID. The compact log-only baseline is a ridge logistic regression over the four features in \autoref{tab:raw-log-feature-defs}. The DAG-derived baseline uses the fixed additive score from Section~\ref{subsec:accuracy-metrics}.

\begin{table}[htbp]
\centering
\small
\setlength{\tabcolsep}{4pt}
\begin{tabular}{lcccccc}
\toprule
Dataset / regime & Acc. & Best single log-only & Compact log-only & DAG-derived & $\Delta$ AUC & $\Delta$ AP \\
\midrule
BrowseComp / sequential        & 0.304 & 0.738 / 0.535 & 0.775 / 0.546 & 0.855 / 0.617 & +0.080 & +0.071 \\
WebWalker-Hard / parallel      & 0.714 & 0.719 / 0.808 & 0.737 / 0.827 & 0.840 / 0.939 & +0.103 & +0.112 \\
DeepSearchQA / sequential      & 0.424 & 0.518 / 0.471 & 0.553 / 0.509 & 0.844 / 0.809 & +0.291 & +0.301 \\
DeepSearchQA / parallel        & 0.336 & 0.593 / 0.460 & 0.541 / 0.412 & 0.856 / 0.726 & +0.315 & +0.313 \\
All sequential                 & 0.321 & 0.628 / 0.503 & 0.664 / 0.527 & 0.849 / 0.713 & +0.186 & +0.186 \\
All parallel                   & 0.615 & 0.656 / 0.634 & 0.639 / 0.620 & 0.848 / 0.832 & +0.209 & +0.212 \\
\bottomrule
\end{tabular}
\caption{Raw-log process baselines versus the fixed DAG-derived score. Each score is AUC / AP under question-held-out evaluation. ``Best single log-only'' reports the strongest individual raw-log feature in each split, while ``Compact log-only'' combines all four raw-log features. The all-regime rows are benchmark-macro summaries.}
\label{tab:raw-log-baseline}
\end{table}

\FloatBarrier
\end{document}